\documentclass[pdflatex,sn-mathphys-num]{sn-jnl}

\usepackage{graphicx}
\usepackage{multirow}
\usepackage{amsmath,amssymb,amsfonts}
\usepackage{amsthm}
\usepackage{mathrsfs}
\usepackage[title]{appendix}
\usepackage{xcolor}
\usepackage{textcomp}
\usepackage{manyfoot}
\usepackage{booktabs}
\usepackage{algorithm}
\usepackage{algorithmicx}
\usepackage{algpseudocode}
\usepackage{listings}
\usepackage{array}
\usepackage{pifont}

\definecolor{checkgreen}{RGB}{0,128,0}
\definecolor{crossred}{RGB}{190,0,0}

\newcommand{\cmark}{\textcolor{checkgreen}{\ding{51}}}
\newcommand{\xmark}{\textcolor{crossred}{\ding{55}}}

\newcolumntype{C}[1]{>{\centering\arraybackslash}m{#1}}

\begin{document}

\title[TWINS]{TWINS: A Tactile Wearable Isomorphic Arm Networked System for Contact-Rich Manipulation Learning}

\author*[1]{\fnm{Takahide} \sur{Kitamura}}
\equalcont{Equal contribution.}

\author[1]{\fnm{Masaki} \sur{Murooka}}
\equalcont{Equal contribution.}

\author[1]{\fnm{Natsuki} \sur{Yamanobe}}

\author[1]{\fnm{Yukiyasu} \sur{Domae}}

\affil*[1]{\orgname{National Institute of Advanced Industrial Science and Technology
  (AIST)}, \orgaddress{\street{2-3-26 Aomi}, \city{Koto-ku},
  \postcode{135-0064}, \state{Tokyo}, \country{Japan}}}

\abstract{
Recent advances in robot learning for manipulation have increased the importance of collecting real-world demonstration data.
However, existing robotic systems primarily focus on end-effector manipulation, making it difficult to teach and execute manipulation tasks involving body-surface contact with the arms and chest.
This paper presents TWINS (Tactile Wearable Isomorphic Arm Networked System), a robotic system for manipulation involving body-surface contact.
TWINS consists of a Wearable Dual-Arm Device, which is worn by the operator, and an Isomorphic Robot with the same joint configuration and external dimensions.
Distributed tactile sensors embedded in the chest and arms enable the measurement of body-surface contact synchronized with joint motion.
Using the Wearable Dual-Arm Device, we collected demonstrations for four manipulation tasks involving body-surface contact.
We then trained imitation learning policies using the collected demonstrations and deployed them on the Isomorphic Robot, enabling manipulation guided by body-surface tactile observations.
Experimental results demonstrate that TWINS provides a unified robotic system for demonstration, learning, and execution of manipulation involving body-surface contact.
Project page: \url{https://mmurooka.github.io/twins-project-page/}

}


\maketitle

\section{Introduction}

\begin{figure}[t]
\centering
\begin{minipage}[t]{0.48\textwidth}
\centering
\includegraphics[width=\linewidth]{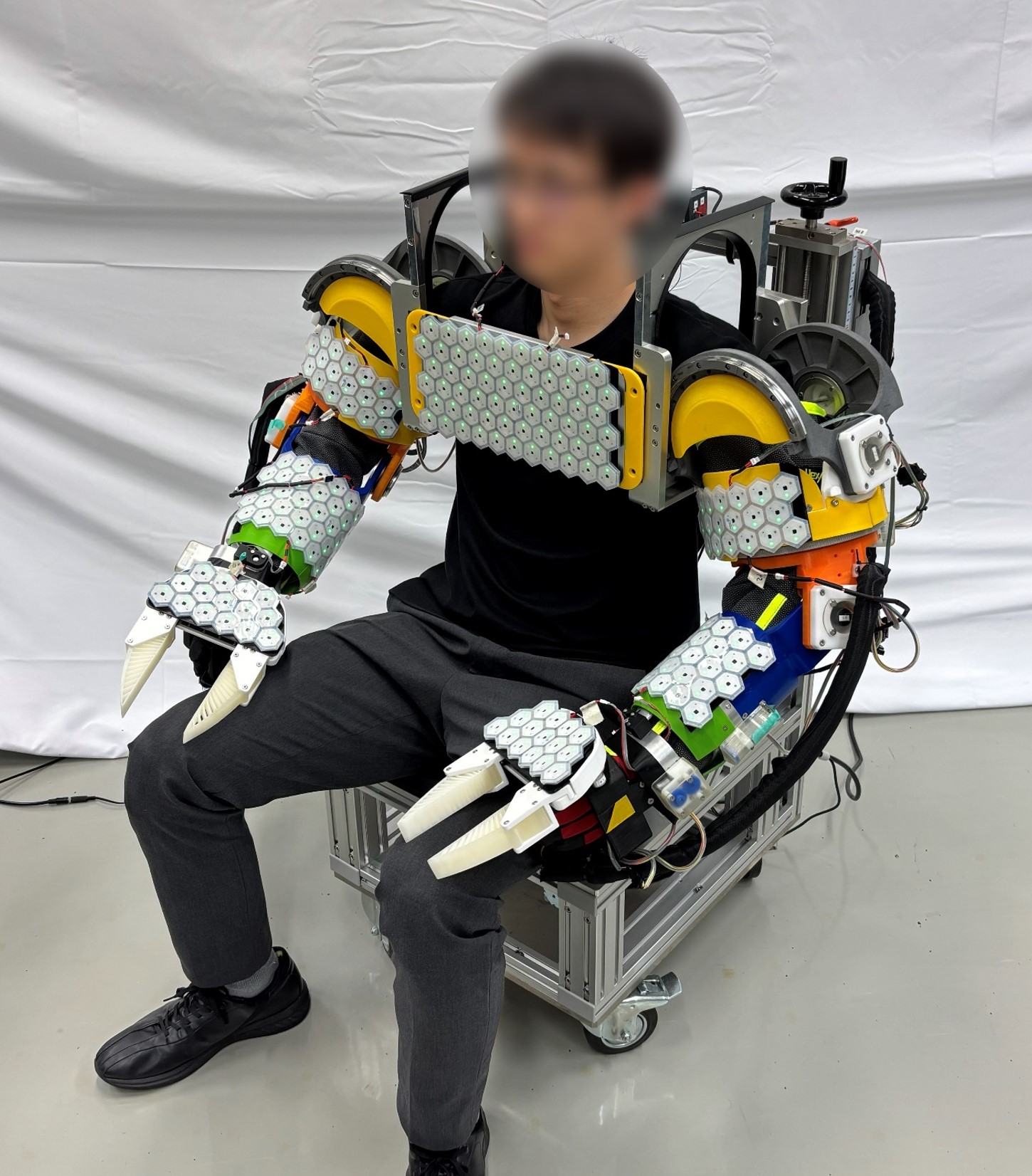}
\par\smallskip \scriptsize
(a) Wearable Dual-Arm Device
\end{minipage}
\hfill
\begin{minipage}[t]{0.48\textwidth}
\centering
\includegraphics[width=\linewidth]{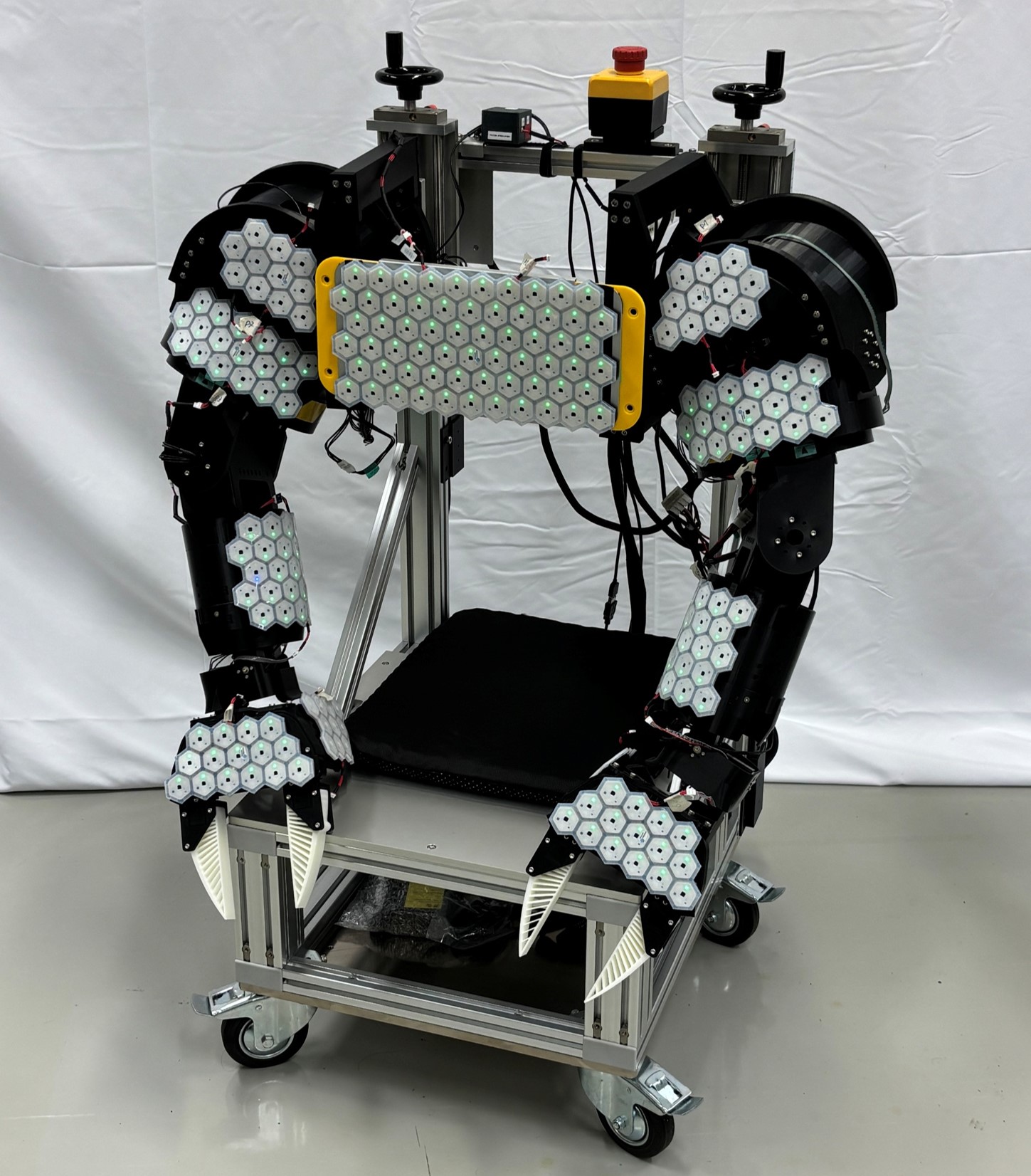}
\par\smallskip \scriptsize
(b) Isomorphic Robot
\end{minipage}
\vspace{4mm}
\caption{TWINS: A tactile wearable isomorphic arm networked system.}
\label{fig:twins-system}
\end{figure}

Recent advances in robot learning have increased interest in leveraging real-world data to enable robots to perform complex tasks in real-world environments~\cite{EmbodiedIntelligenceReview:Zhang:FRAI2025}.
In particular, imitation learning from human demonstrations has emerged as one of the most promising approaches for humanoid manipulation~\cite{ILSurvey:Osa:FTR2018}.
While imitation learning enables robots to acquire task-specific skills directly from human demonstrations, its performance largely depends on the quality of the collected demonstrations and the available sensory observations~\cite{RH20T:Fang:ICRA2024}.
Therefore, in addition to developing more advanced learning algorithms, an equally important challenge is to collect real-world demonstrations that faithfully capture the robot's interaction with its environment~\cite{OpenX:OpenX:ICRA2024}.

Physical contact between the robot, manipulated objects, and the surrounding environment plays a crucial role in many manipulation tasks~\cite{ConRichSurvey:Tsuji:IJRR2025}.
Unlike conventional manipulators, humanoid robots can utilize not only their end effectors but also body surfaces, such as the forearms, upper arms, shoulders, and chest, to support or constrain objects during manipulation.
For example, carrying a large object by embracing it with the chest and both arms or transporting an object hooked over an arm requires not only coordinated joint motion but also appropriate body-surface contact.
However, accurately estimating such contact information from vision alone remains challenging~\cite{TACT:Murooka:RAL2025}, and the lack of explicit contact observations may limit the range of manipulation skills that can be learned.



Teleoperation is one of the most widely used approaches for collecting real-world demonstrations~\cite{TeleopReview:Liang:Neuro2025}.
In particular, Leader--Follower systems, which provide corresponding mechanisms for the operator and the robot, are well suited for collecting robot learning data because the robot directly follows the joint motions of the operator-side mechanism~\cite{HumanoidExo:Zhong:arXiv2025}.
However, most existing systems are primarily designed for teaching end-effector poses and grasping motions~\cite{ALOHA:Zhao:RSS2023}.
As a result, operators cannot physically interact with objects using a mechanism equivalent to the robot's body, making it difficult to collect demonstrations involving body-surface contact.
Moreover, when the operator-side device and the robot differ in body dimensions or joint configuration, additional mapping is required to transfer human motions and contact information to the robot.

To collect demonstrations involving body-surface contact, it is necessary not only to measure the operator's motions but also to allow the operator to physically interact with objects and the environment through a body morphology corresponding to that of the robot.
If the operator directly wears a mechanism with the same or a similar joint configuration and external dimensions as the robot, the demonstrated joint motions can be transferred directly to the execution robot while body-surface contact can be measured during demonstration collection.
Such morphological correspondence is expected to reduce structural discrepancies between demonstration collection and robot execution, thereby facilitating the learning of contact-rich manipulation.



To address these challenges, this paper presents TWINS (Tactile Wearable Isomorphic Arm Networked System), a robotic system for imitation learning of manipulation involving body-surface contact.
An overview of the proposed system is shown in Figure~\ref{fig:twins-system}.
The system consists of a Wearable Dual-Arm Device worn by the operator and a corresponding Isomorphic Robot.
Both systems share the same joint configuration and external dimensions, reducing structural discrepancies between demonstration collection and robot execution.
By manipulating objects through the Wearable Dual-Arm Device, the operator can physically experience body-surface contact while joint motions and tactile observations are synchronously recorded.

This paper presents the mechanical design, sensor configuration, control system, and robot learning pipeline of TWINS.
Demonstrations are collected using the Wearable Dual-Arm Device for multiple manipulation tasks involving body-surface contact, and imitation learning policies trained on the collected demonstrations are deployed on the Isomorphic Robot.
Experimental results demonstrate that the proposed system enables the collection of manipulation data containing body-surface contact with the arms and chest, which is difficult to obtain using conventional end-effector-centric demonstration systems, and effectively leverages such data for robot learning.



\section{Related Works}

\begin{table}[t]
\caption{Comparison of robot demonstration interfaces.}
\label{tab:comparison}
\centering
\fontsize{9pt}{10.8pt}\selectfont
\setlength{\tabcolsep}{4pt}
\renewcommand{\arraystretch}{1.15}

\begin{tabular}{
>{\raggedright\arraybackslash}m{0.38\columnwidth}
C{0.16\columnwidth}
C{0.18\columnwidth}
C{0.18\columnwidth}
}
\toprule

\textbf{Method}
&
\shortstack{\textbf{Embodied}\\\textbf{Operation}}
&
\shortstack{\textbf{Morphological}\\\textbf{Correspondence}}
&
\shortstack{\textbf{Body-Surface}\\\textbf{Contact}}
\\
\midrule

Motion Capture + Tactile Suit~\cite{ActionSense:DelPreto:NeurIPS2022}
&
\cmark
&
\xmark
&
\cmark
\\

VR Teleoperation~\cite{OmniH2O:He:CoRL2024}
&
\cmark
&
\xmark
&
\cmark
\\

Leader--Follower Robot~\cite{CHILD:Myers:Humanoids2025}
&
\xmark
&
\cmark
&
\cmark
\\

Wearable End-Effector~\cite{UMI:Chi:RSS2024}
&
\cmark
&
\cmark
&
\xmark
\\

Exoskeleton~\cite{MasterRobot:Kim:RAL2023}
&
\cmark
&
\cmark
&
\xmark
\\

\midrule

\textbf{TWINS (Proposed)}
&
\cmark
&
\cmark
&
\cmark
\\

\bottomrule
\end{tabular}
\end{table}

This section reviews robot demonstration interfaces related to this study.

Motion capture has been widely used to directly measure human body motion for robot demonstrations~\cite{HumanoidRetarget:Ayusawa:TRO2017,TeleopHumanoid:Penco:RAM2019}.
More recently, systems such as ActionSense have combined motion capture with wearable tactile sensors to simultaneously acquire body motion and body-surface contact during demonstrations~\cite{ActionSense:DelPreto:NeurIPS2022}.
While these approaches can capture the operator's motion and contact information, they do not account for the structural differences between the human body and the robot during demonstration collection.

VR teleoperation has also been extensively studied as an intuitive interface for robot demonstration.
For example, OmniH2O enables intuitive whole-body teleoperation of a humanoid robot using a VR device~\cite{OmniH2O:He:CoRL2024}.
Although VR teleoperation provides an immersive operation experience, operators cannot physically interact with objects or the environment, making it difficult to obtain tactile feedback and body-surface contact information~\cite{TACT:Murooka:RAL2025}.




Leader--Follower Robots, which use corresponding robots for the operator and the execution robot, are also widely used for robot demonstration~\cite{MEVION:Kawaharazuka:arXiv2026}.
Representative systems such as ALOHA~\cite{ALOHA:Zhao:RSS2023,MobileALOHA:Fu:CoRL2024} can synchronously collect multimodal data, including joint angles and visual observations.
Furthermore, CHILD extends the Leader--Follower Robot paradigm to humanoid robots, enabling whole-body manipulation~\cite{CHILD:Myers:Humanoids2025}.
However, feedback of the Follower Robot's contact state to the operator is limited, making it difficult for the operator to physically experience body-surface contact during demonstration collection.

Wearable End-Effectors, in which the operator directly wears the robot's end effector, have also been proposed.
UMI enables demonstration collection using the same end effector as the robot, allowing demonstrations that are not constrained by the robot arm's workspace~\cite{UMI:Chi:RSS2024}.
Because the same end effector is used during both demonstration collection and execution, grasping motions and contact states can be naturally transferred to the robot.
However, the correspondence is limited to the end effector and does not support body-surface contact involving the arms or chest.



Exoskeleton-based interfaces have also been proposed to cover a wider range of body parts~\cite{MasterRobot:Kim:RAL2023,TABLIS:Ishiguro:RAL2020}.
By using an exoskeleton, human joint motions can be accurately measured and directly transferred to a corresponding robot.
More recently, humanoid exoskeletons such as HumanoidExo and HOMIE have been developed, demonstrating the potential for large-scale demonstration collection~\cite{HumanoidExo:Zhong:arXiv2025,HOMIE:Ben:RSS2025}.
However, these systems focus primarily on capturing body motion rather than acquiring body-surface contact corresponding to the robot.

This study compares existing robot demonstration interfaces from three perspectives: Embodied Operation, Morphological Correspondence, and Body-Surface Contact.
Embodied Operation refers to the ability to intuitively operate the robot from a first-person perspective, allowing the operator to feel as if they were embodied in the robot.
Morphological Correspondence refers to the structural similarity between the demonstration interface and the execution robot, enabling demonstrated motions to be transferred with little or no transformation.
Body-Surface Contact refers to the ability to acquire contact information between the robot body, including the arms and chest, and manipulated objects or the environment.
As summarized in Table~\ref{tab:comparison}, existing systems satisfy some of these characteristics, but none achieves all three simultaneously.
In contrast, the proposed TWINS integrates all three by combining a Wearable Dual-Arm Device with an Isomorphic Robot.



\section{Design of TWINS}

\subsection{Mechanical Design}

\begin{figure}[t]
\centering
\begin{minipage}[t]{0.38\textwidth}
\centering
\includegraphics[width=\linewidth]{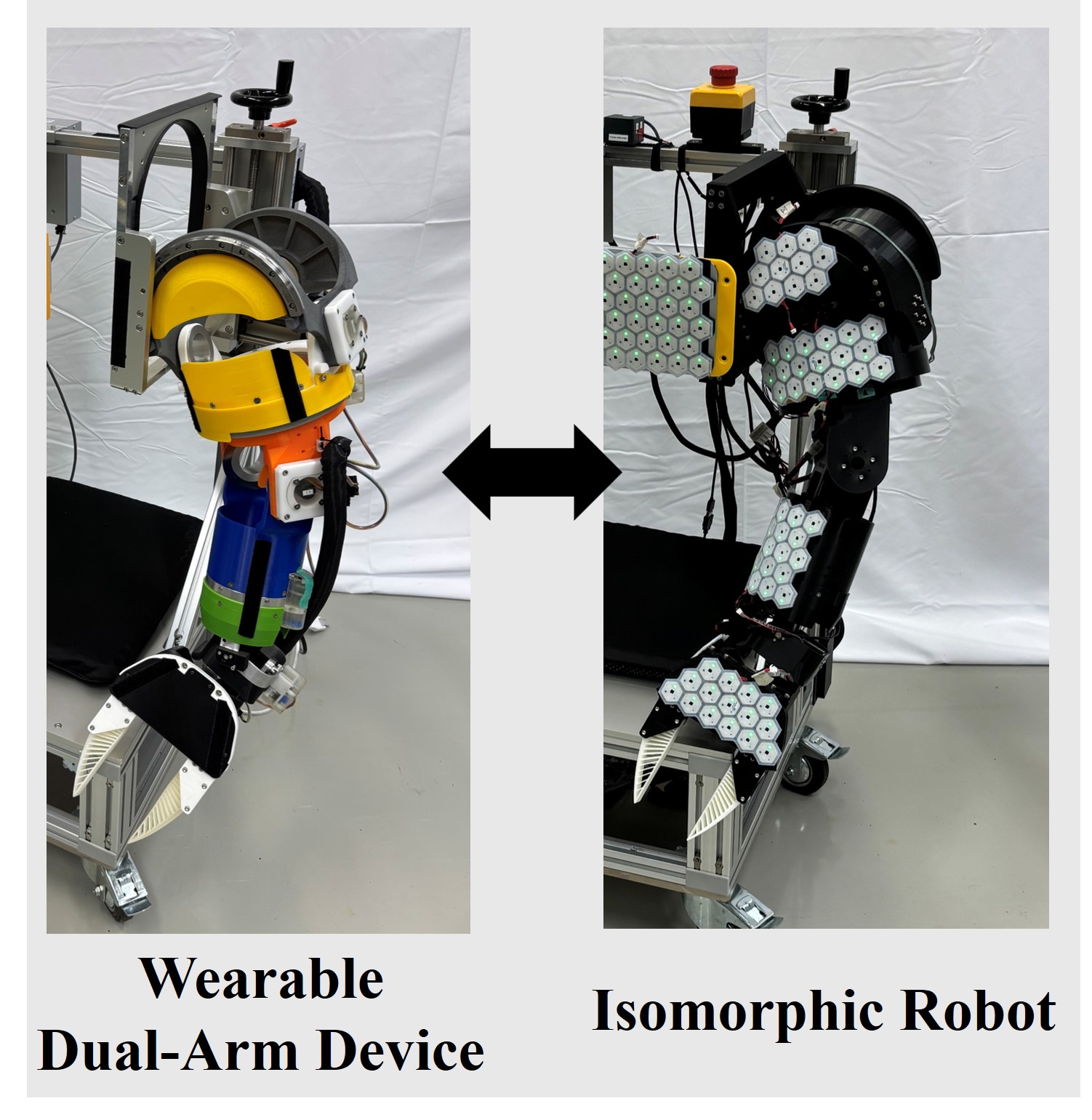}
\par\smallskip \scriptsize
(a) Isomorphic design
\end{minipage}
\hfill
\begin{minipage}[t]{0.306\textwidth}
\centering
\includegraphics[width=\linewidth]{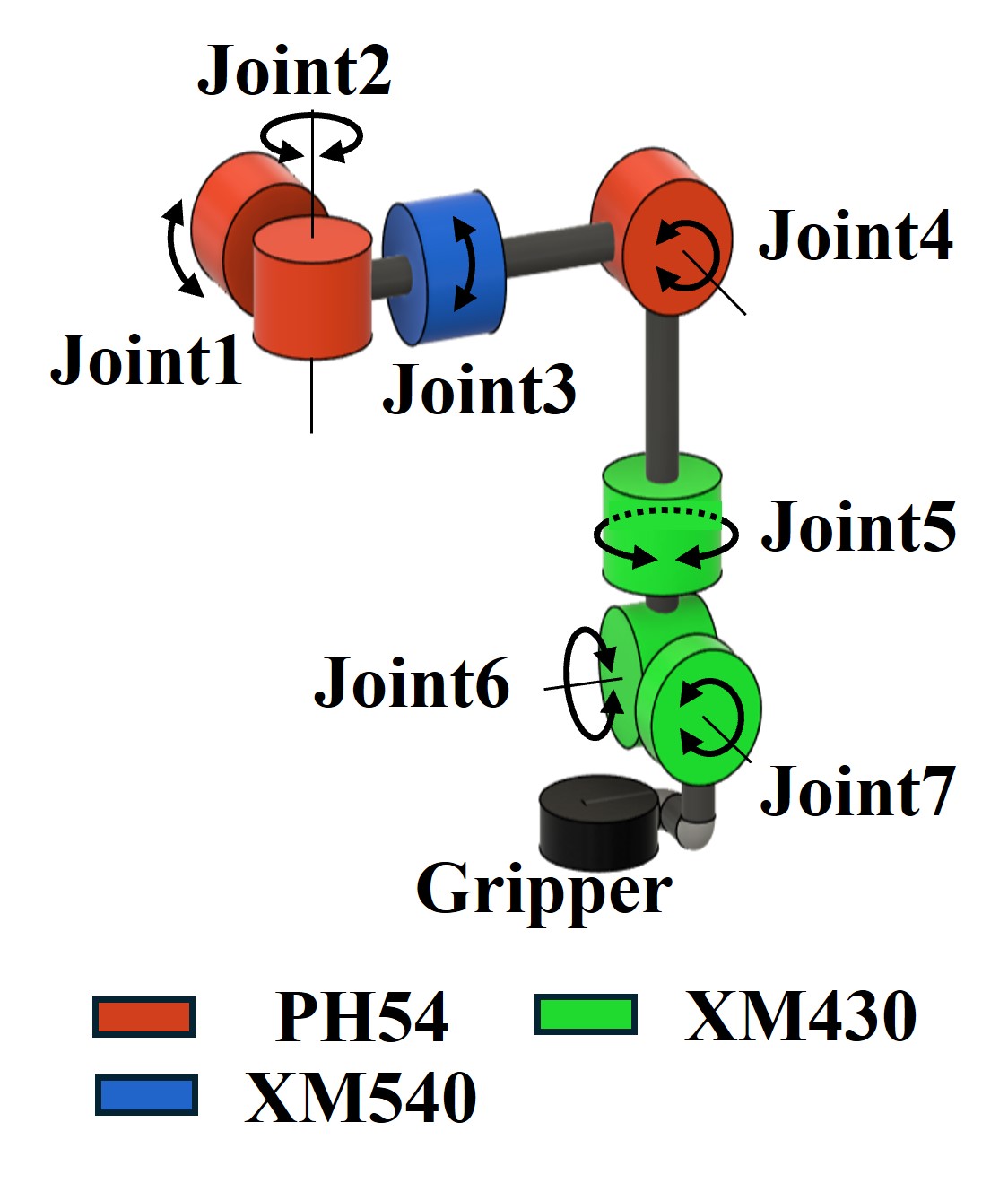}
\par\smallskip \scriptsize
(b) Joint configuration
\end{minipage}
\hfill
\begin{minipage}[t]{0.287\textwidth}
\centering
\includegraphics[width=\linewidth]{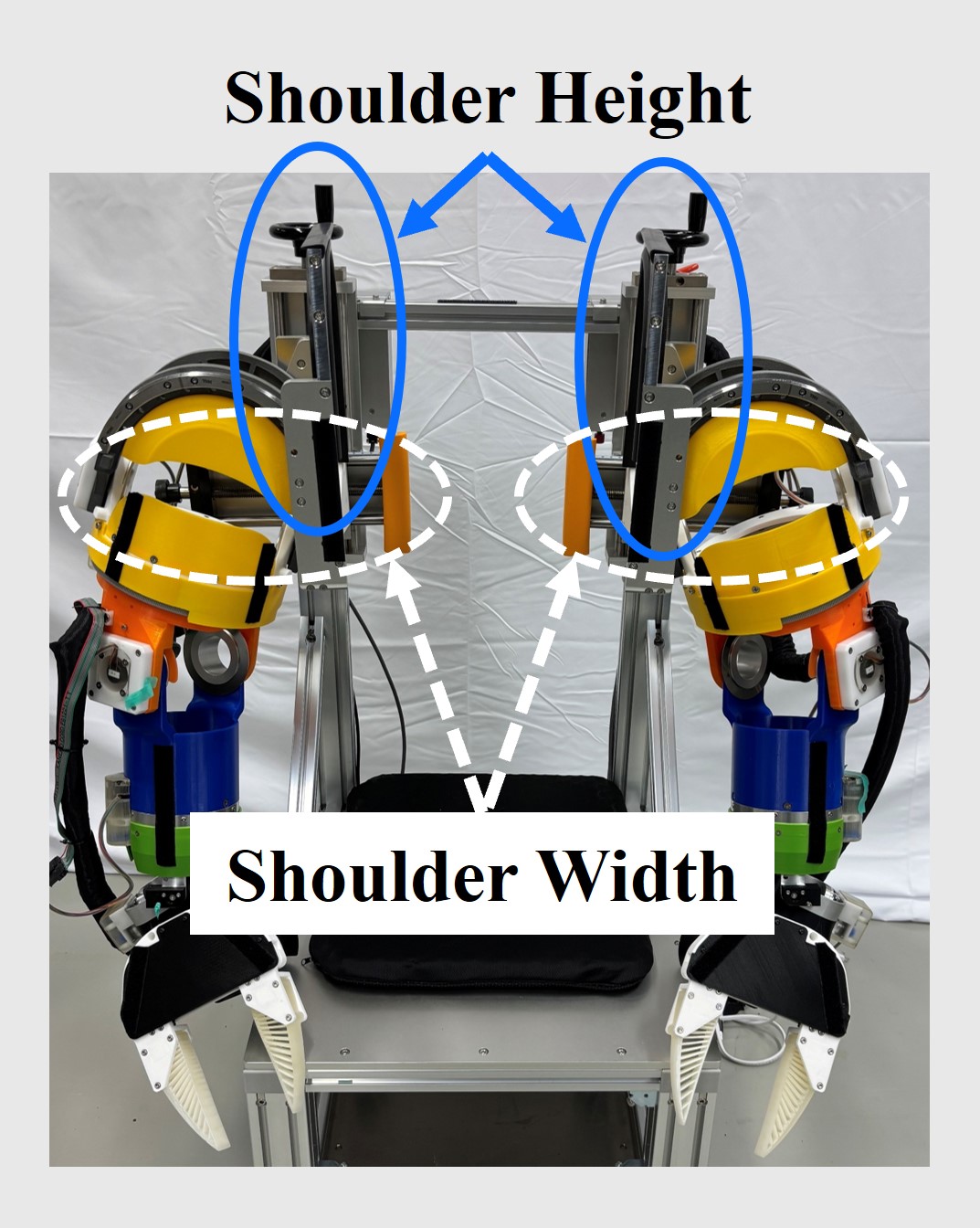}
\par\smallskip \scriptsize
(c) Shoulder adjustment mechanism
\end{minipage}
\vspace{4mm}
\caption{Mechanical design of TWINS.
\newline\footnotesize{
TWINS consists of a Wearable Dual-Arm Device and an Isomorphic Robot that share the same joint configuration and body dimensions. The Wearable Dual-Arm Device includes manual adjustment mechanisms for shoulder width and shoulder height to accommodate operators of different body sizes.
}}
\label{fig:mechanical-design}
\end{figure}

TWINS consists of a Wearable Dual-Arm Device worn by the operator to collect demonstrations and an Isomorphic Robot that executes motions based on the collected demonstrations.
Figure~\ref{fig:mechanical-design}(a) shows the appearance and mechanical structure of the left arm.
The Wearable Dual-Arm Device and the Isomorphic Robot share the same joint configuration and external dimensions, allowing joint motions and body-surface contact acquired through the Wearable Dual-Arm Device to be directly transferred to the Isomorphic Robot.

Each arm has seven degrees of freedom, comprising a three-degree-of-freedom shoulder, a one-degree-of-freedom elbow, and a three-degree-of-freedom wrist.
Figure~\ref{fig:mechanical-design}(b) illustrates the joint configuration.
Rather than faithfully reproducing the full range of motion of the human upper limb, the arm was designed based on the joint configuration commonly adopted in existing humanoid robots~\cite{HRP2Kai:Kaneko:Humanoids2015}.
The link lengths were determined based on typical adult upper-limb dimensions so that the Wearable Dual-Arm Device can be worn in a natural posture.
For the end effector, we adopted a parallel two-finger gripper with a Fin Ray structure designed based on UMI~\cite{UMI:Chi:RSS2024}.



The Wearable Dual-Arm Device has a hollow structure that allows the operator's arms to be inserted into each arm.
Because the operator's arms are enclosed by an outer shell, including the joint sections, objects and the environment are manipulated through the shell corresponding to the Isomorphic Robot without direct physical contact by the operator.
This design enables the reproduction of not only the operator's joint motions but also body-surface contact occurring on the outer shell, which has the same geometry as the Isomorphic Robot.
Each joint is equipped with a 10-bit encoder for measuring joint angles.

To accommodate operators with different body sizes, the Wearable Dual-Arm Device provides independent adjustment mechanisms for shoulder width and the height of each shoulder.
As shown in Figure~\ref{fig:mechanical-design}(c), the base supporting each arm is mounted on linear guides with manual adjustment mechanisms, allowing the arm positions to be aligned with the operator's shoulders.
Furthermore, to eliminate the need for the operator to support the weight of both arms, the shoulder support structure is fixed to a chair so that the weight of the entire device is supported by the chair.

The Isomorphic Robot shares the same joint configuration, link lengths, and external dimensions as the Wearable Dual-Arm Device.
This isomorphic design allows the joint angles measured by the Wearable Dual-Arm Device to be directly transferred to the Isomorphic Robot without complex retargeting.
As a result, transformation errors caused by differences in body morphology are minimized, ensuring faithful reproduction of the demonstrated motions.




Each joint of the Isomorphic Robot is actuated by a DYNAMIXEL-series servo motor.
XM430, XM540, and PH54-100-S500 actuators were selected according to the required joint torque and link weight.
High-torque actuators were used for joints subjected to large moments, such as the shoulder, and multiple actuators were employed where necessary to provide sufficient driving torque.
In contrast, lightweight actuators were used for distal joints, such as the wrist, to reduce both the weight of the distal links and the loads acting on the joints.

TWINS will be released as open-source hardware to facilitate future research.
The released resources will include a bill of materials, 3D-printable models, and control software.
Most mechanical components, including the outer shells and links, were fabricated using 3D printing.
This approach provides the lightweight structure required for a wearable device while also simplifying design modifications and part replacement, making the system well suited for reproduction and continuous improvement.



\subsection{Tactile Sensing System}

\begin{figure}[t]
\centering
\includegraphics[height=0.30\textwidth]{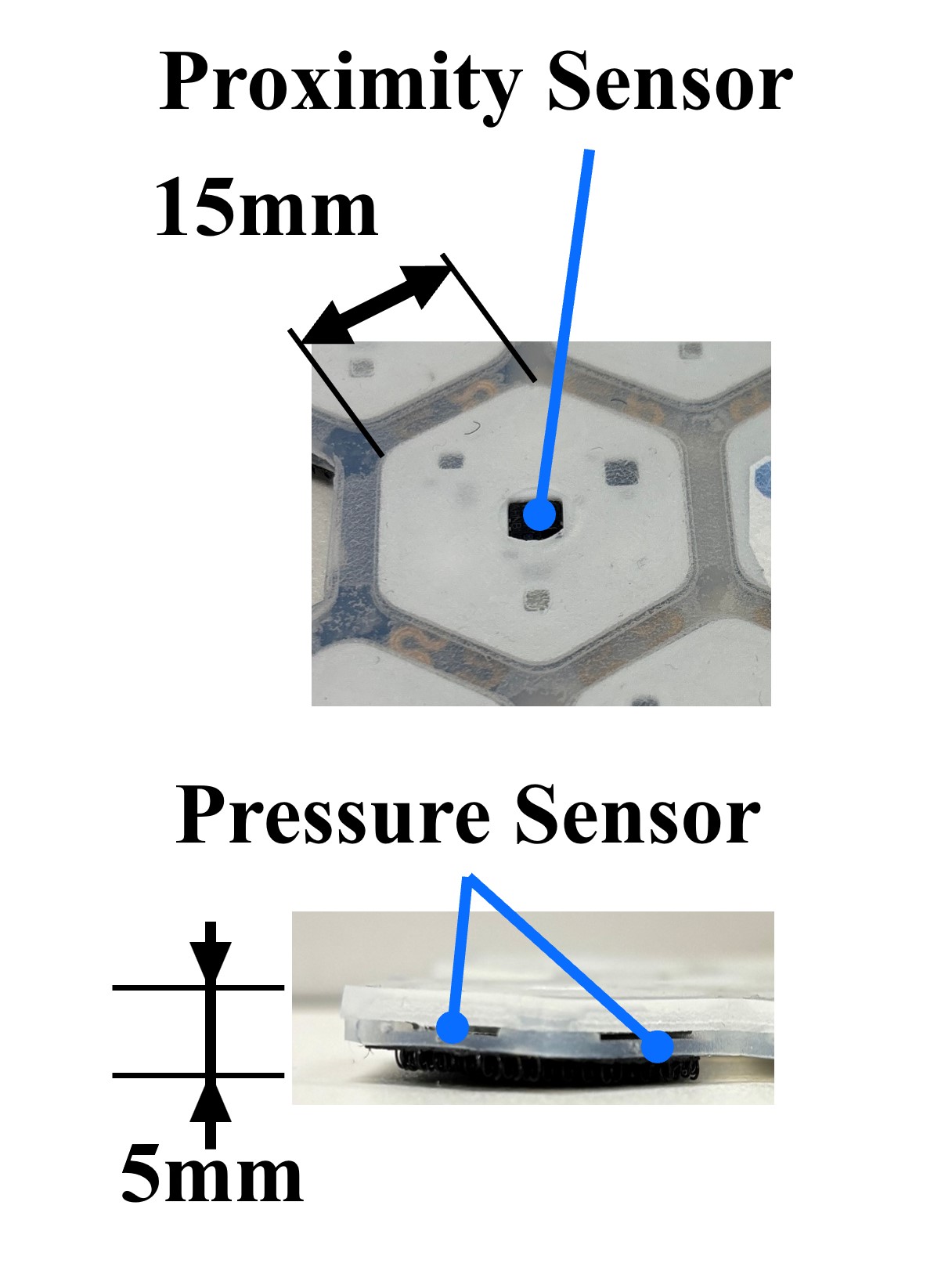}
\hfill
\includegraphics[height=0.30\textwidth]{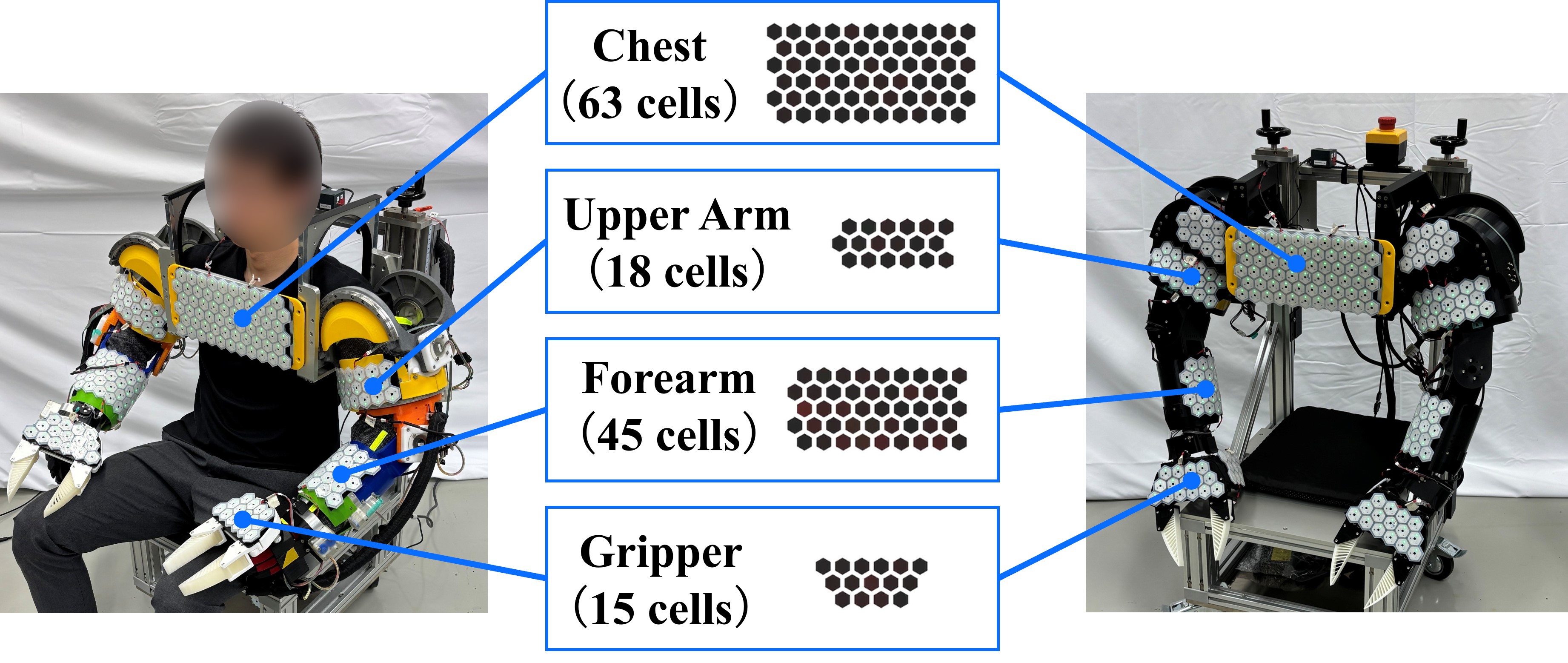}
\par\smallskip
\begin{minipage}[t]{0.24\textwidth}
\centering\scriptsize
(a) e-Skin cell
\end{minipage}
\hfill
\begin{minipage}[t]{0.72\textwidth}
\centering\scriptsize
(b) e-Skin layout on TWINS
\end{minipage}
\vspace{4mm}
\caption{
Tactile sensing system.
\newline\footnotesize{
(a) Each hexagonal e-Skin cell measures both pressure and proximity.
(b) Tactile sensors are mounted on the chest, front of the upper arm, inner forearm, and top surface of the gripper on the Wearable Dual-Arm Device and the Isomorphic Robot to capture body-surface contact during manipulation tasks.}}
\label{fig:eskin}
\end{figure}

TWINS uses e-Skin tactile sensors manufactured by Intouch Robotics~\cite{RobotSkin:Cheng:IEEE2019} to measure contact with objects and the environment on the outer surfaces of the arms and chest.
Figure~\ref{fig:eskin}(a) shows the appearance of the sensor.
The sensor has a sheet-like structure composed of regular hexagonal cells connected by a silicone material.
Each cell has a side length of 15~mm and a thickness of approximately 5~mm and can measure both pressure and proximity.

The tactile sensors are arranged as patches consisting of multiple cells at the expected contact regions, as shown in Figure~\ref{fig:eskin}(b).
Each arm contains 15 cells on the upper surface of the gripper, 45 cells on the inner surface of the forearm, and 18 cells on the front surface of the upper arm, while 63 cells are placed on the chest.
The sensor arrangement on the arms is bilaterally symmetric.
This arrangement enables the measurement of manipulation involving distributed contact not only at the grippers but also over the forearms, upper arms, and chest.



Each sensor patch is detachable and can be mounted at corresponding locations on both the Wearable Dual-Arm Device and the Isomorphic Robot.
In the Wearable Dual-Arm Device, the tactile sensors are attached to the outer shell rather than directly to the operator's skin.
Because the shell has the same geometry as the Isomorphic Robot, the same sensor layout can be used during both demonstration collection and robot execution.
Consequently, the contact information acquired by the Wearable Dual-Arm Device can be directly interpreted as body-surface contact on the Isomorphic Robot.

Sensor data are acquired at 10~Hz.
Rather than representing contact as a single point, the tactile information at each body region is treated as a spatial distribution over multiple sensor cells.
This representation captures not only the presence of contact but also the extent of the contact region and the pressure distribution.
Furthermore, the proximity sensing capability of each cell enables observation of object distance before physical contact, allowing continuous capture of the contact process from approach to separation.



\subsection{Data Collection and Motion Execution}

TWINS synchronously records joint motions and body-surface contact using the Wearable Dual-Arm Device for subsequent execution on the Isomorphic Robot.
The state at time $t$ is defined as
\begin{equation}
\mathbf{x}(t)=
\left\{
\mathbf{q}(t),
\mathbf{S}(t)
\right\},
\label{eq:state}
\end{equation}
where $\mathbf{q}(t)$ denotes the joint angles of the Wearable Dual-Arm Device and $\mathbf{S}(t)$ represents the pressure and proximity measurements from all tactile sensor cells.
A demonstration is recorded as the time series
\begin{equation}
\mathcal{D}
=
\{
\mathbf{x}(t)
\}_{t=0}^{T},
\label{eq:dataset}
\end{equation}
where $T$ is the demonstration length.

Because the Wearable Dual-Arm Device and the Isomorphic Robot share the same joint configuration and link lengths, the joint angles can be directly transferred without retargeting.
The joint angles of the Isomorphic Robot are given by
\begin{equation}
\mathbf{q}_{\mathrm{R}}(t)
=
\mathbf{q}_{\mathrm{W}}(t),
\end{equation}
where $\mathbf{q}_{\mathrm{W}}(t)$ denotes the joint angles of the Wearable Dual-Arm Device.
This direct mapping enables faithful reproduction of the demonstrated motions without kinematic transformation.

Demonstrations are collected by having the operator wear the Wearable Dual-Arm Device and directly manipulate objects.
The recorded joint trajectories can be replayed on the Isomorphic Robot for system validation.
Furthermore, the recorded time series, which integrates joint motions and body-surface contact, can be used for imitation learning of contact-rich manipulation.




\section{Experiments}

\subsection{Contact-Rich Manipulation Tasks}

TWINS was evaluated on four representative manipulation tasks involving body-surface contact.
In each task, the manipulation proceeds through multiple phases according to body-surface contact events.
To enable repeated execution, every task starts from a predefined idle pose and returns to the same pose after completion.
The manipulation phases of each task are described below, corresponding to the numbered phases in Figure~\ref{fig:teleop}.


\begin{itemize}

\item \textbf{Towel Hanging}: A task in which the robot changes its posture according to a towel placed on its arm.
(1) The robot starts from the idle pose.
(2) An assistant places a towel over either the left or right forearm.
(3) The arm holding the towel rotates outward while supporting it.
(4) After the towel is removed, (5) the robot returns to the idle pose.
The side, order, and timing of towel presentation are randomized across trials.
The towel was folded to 70 $\times$ 35 $\times$ 2~cm and presented to the robot.

\item \textbf{Basket Holding}: A task in which the robot holds a basket with both arms.
(1) The robot starts from the idle pose.
(2) When a basket is pressed against the chest, (3) the robot grasps it with both arms.
(4) Contact with the right forearm triggers basket release, after which (5) the robot returns to the idle pose.
The basket measured 34 $\times$ 34 $\times$ 57~cm.

\item \textbf{Ball Placing}: A task in which the robot places balls into a basket.
(1) The robot starts from the idle pose.
(2) The robot holds the basket against its chest with the right arm.
(3) Contact on the top surface of the gripper triggers grasping of a ball presented to the left gripper, followed by placement into the basket.
Ball grasping and placement are repeated in response to subsequent contact events.
Finally, (4) contact with the left forearm triggers basket release, and (5) the robot returns to the idle pose.
The basket measured 24 $\times$ 17 $\times$ 18~cm, and each ball had a diameter of 5~cm.

\item \textbf{Adaptive Holding}: A task in which the robot switches between different holding strategies according to the contact location.
(1) The robot starts from the idle pose.
(2) An assistant presses either a long cardboard box against the chest or a packed sleeping bag against one forearm.
(3) The robot selects an appropriate holding motion according to the contact location.
Specifically, the box is supported with both arms, whereas the packed sleeping bag is held by the contacted arm.
(4) After the object is removed, (5) the robot returns to the idle pose.
The order and timing of object presentation are randomized across trials.
The cardboard box measured 18 $\times$ 16 $\times$ 117~cm and weighed 0.6~kg, while the packed sleeping bag measured 24~cm in diameter and 37~cm in height and weighed 1.3~kg.





\end{itemize}

\subsection{Demonstration Collection}

Demonstration data for the four manipulation tasks were collected using the Wearable Dual-Arm Device of TWINS.
The operator performed demonstrations by wearing the Wearable Dual-Arm Device and directly manipulating the objects.
To ensure consistent demonstrations across trials, all data were collected by a single operator.
For all tasks, an assistant presented and removed the objects and generated the body-surface contact events defined by each task.
To encourage natural manipulation, object positions and orientations were not fixed, and the demonstrations were collected under manually varied presentation conditions.

\begin{figure}[!htbp]
\centering
\includegraphics[width=0.85\textwidth]{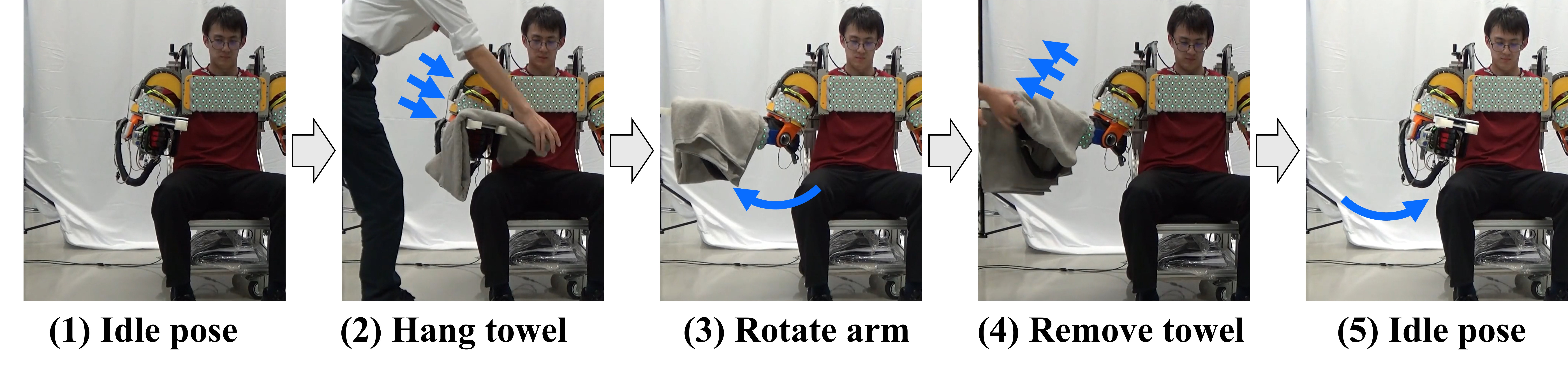}
\par\smallskip \scriptsize
(a) Towel Hanging
\par\medskip
\vspace{2mm}
\includegraphics[width=0.85\textwidth]{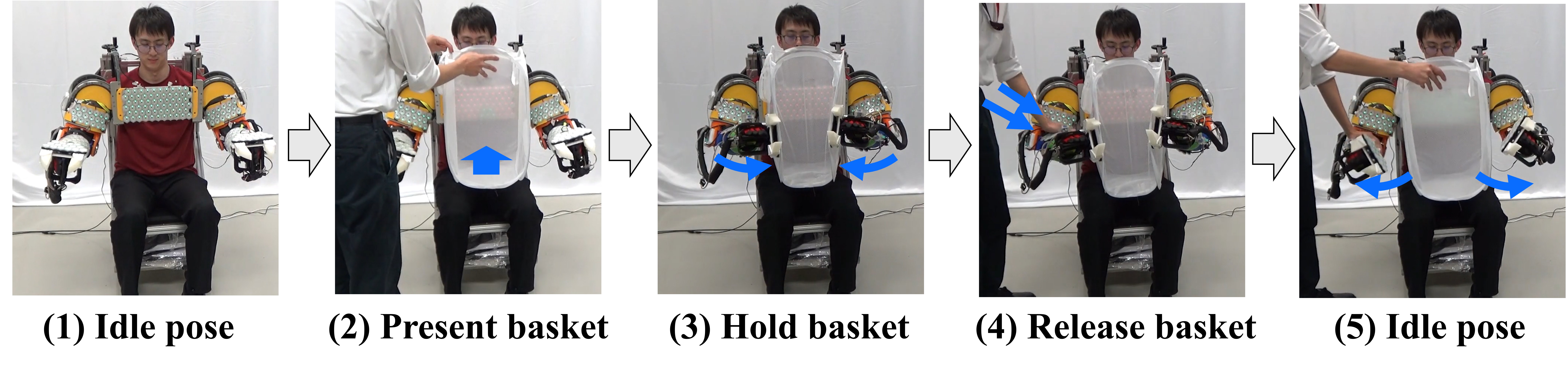}
\par\smallskip \scriptsize
(b) Basket Holding
\par\medskip
\vspace{2mm}
\includegraphics[width=0.85\textwidth]{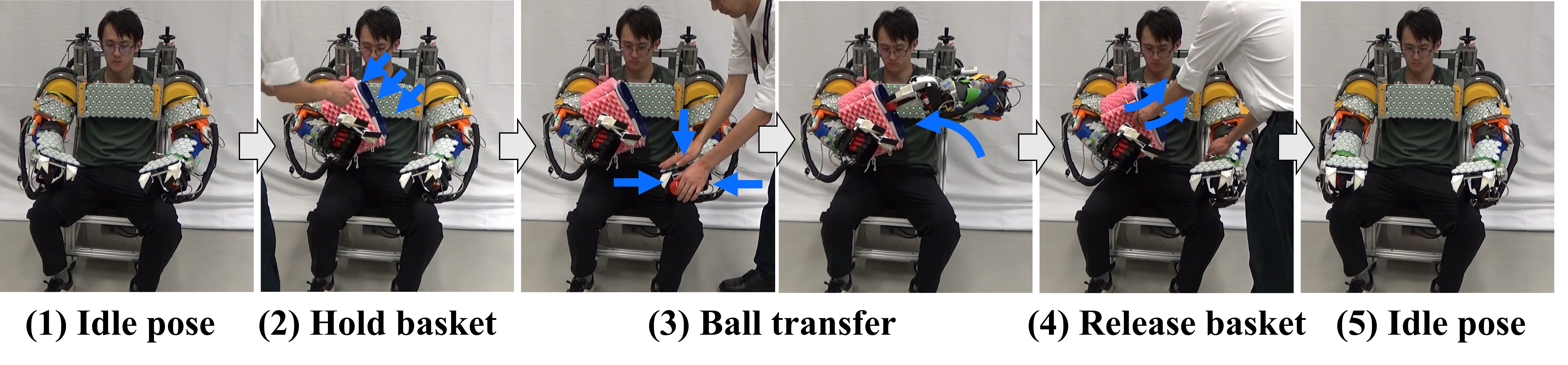}
\par\smallskip \scriptsize
(c) Ball Placing
\par\medskip
\vspace{2mm}
\includegraphics[width=0.85\textwidth]{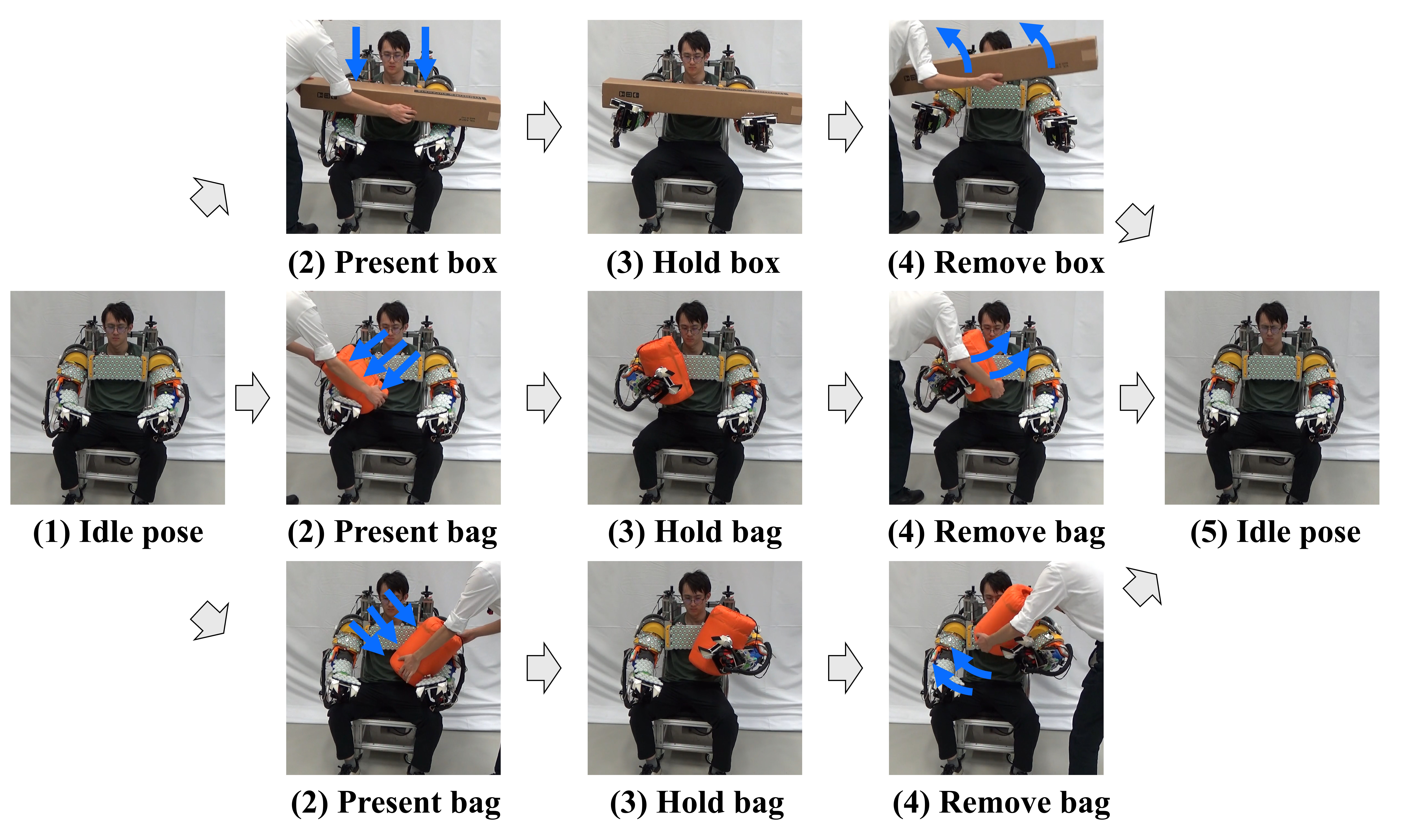}
\par\smallskip \scriptsize
(d) Adaptive Holding
\vspace{4mm}
\caption{Contact-rich demonstration collection using the Wearable Dual-Arm Device.}
\label{fig:teleop}
\end{figure}

\begin{figure}[!htbp]
\centering
\includegraphics[width=0.9\textwidth]{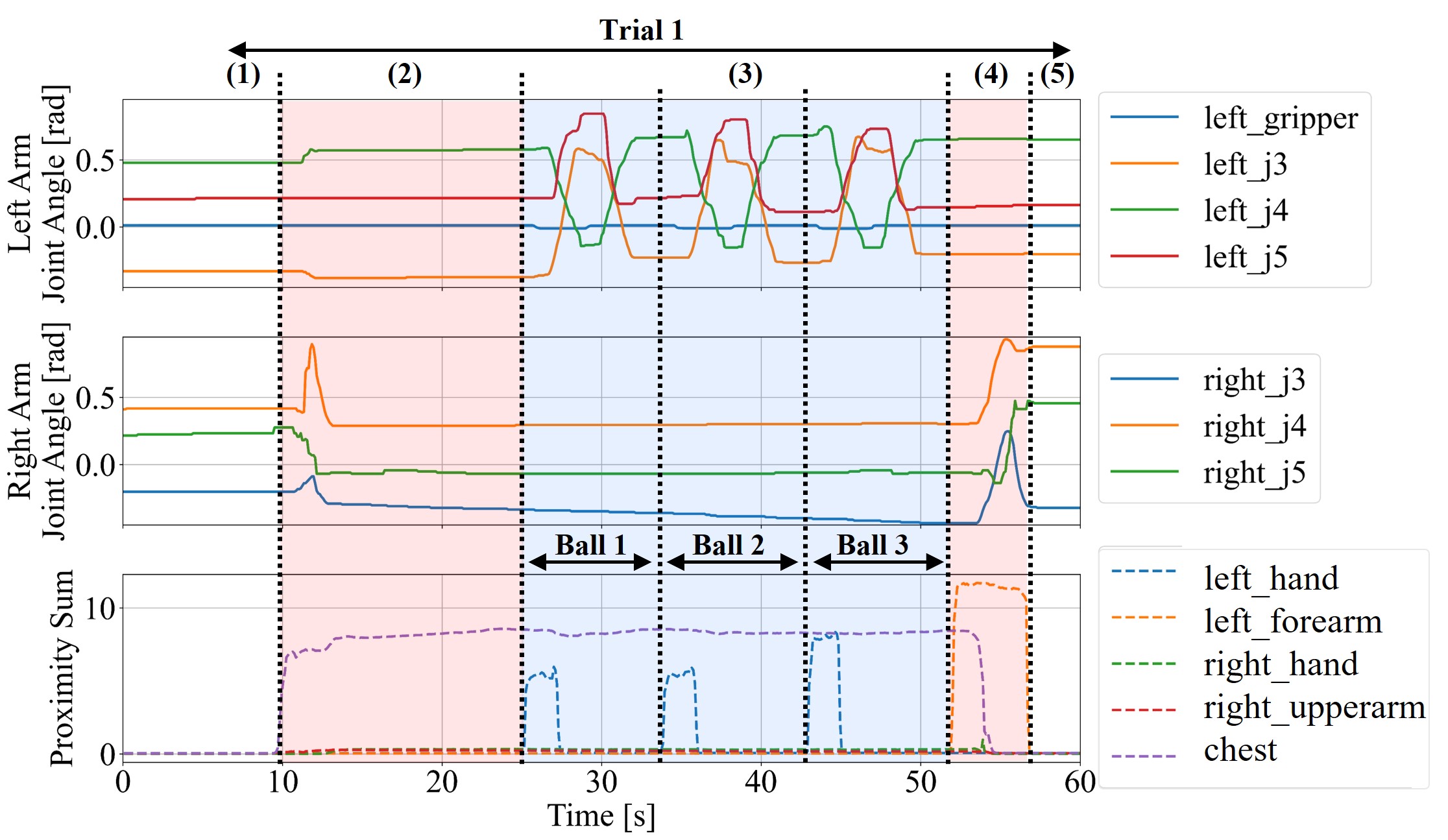}
\par\smallskip \scriptsize
(a) Ball Placing
\par\medskip
\vspace{2mm}
\includegraphics[width=0.9\textwidth]{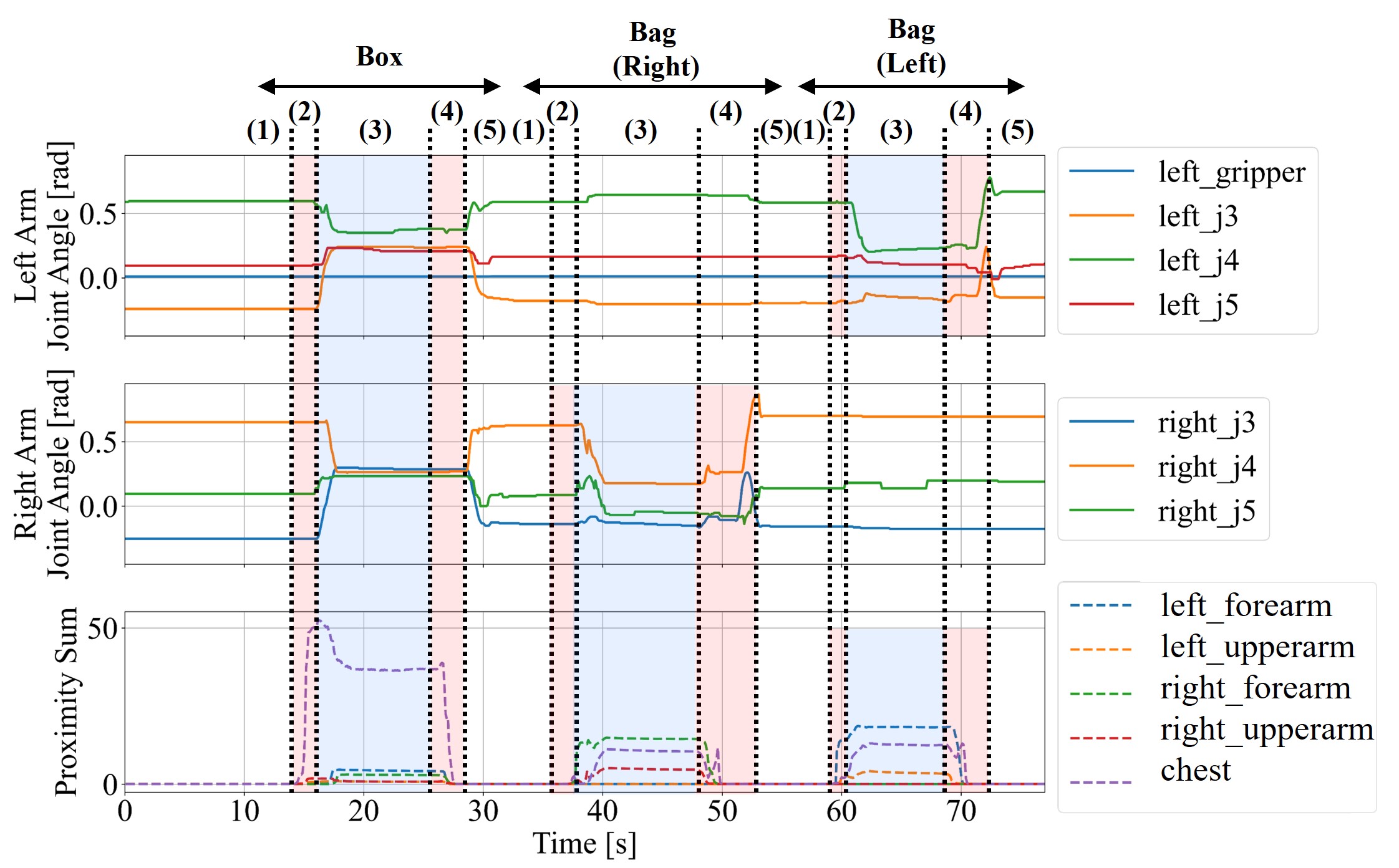}
\par\smallskip \scriptsize
(b) Adaptive Holding
\vspace{4mm}
\caption{Joint-angle and tactile-sensor data collected with the Wearable Dual-Arm Device.
\newline\footnotesize{
The numbers above the plots correspond to the manipulation phases in Figure~\ref{fig:teleop}.
Only representative joint angles and tactile sensor patches are shown for clarity.
For each tactile sensor patch, the plotted value is the summed proximity measurement over all sensor cells.
}}
\label{fig:teleop-graphs}
\end{figure}

Ten demonstrations were collected for each task, resulting in a total of 40 demonstrations.
The joint angles and tactile sensor measurements defined in Eq.~(\ref{eq:state}) were recorded at 10~Hz to construct the time-series dataset defined in Eq.~(\ref{eq:dataset}).
Figure~\ref{fig:teleop} shows demonstrations of the four tasks.
Figure~\ref{fig:teleop-graphs} presents examples of the joint-angle and tactile sensor data collected during the Ball Placing and Adaptive Holding tasks.
The plots show that the joint motions transition according to body-surface contact events.

The operator was able to perform all demonstrations without requiring specialized training.
Data collection for 10 demonstrations of each task was completed within 30~min, demonstrating that TWINS enables efficient collection of real-world demonstration data.




\subsection{Imitation Learning Setup}

The collected demonstration data were used to train a policy that predicts the next action from the current state through imitation learning.

The state consists of the joint angles and body-surface tactile measurements.
The joint-angle vector has 16 dimensions, corresponding to the left and right arms and grippers.
The tactile input has 438 dimensions, obtained by concatenating the pressure and proximity measurements from 219 tactile sensor cells.
The action is defined as the joint angles at the next time step.

Diffusion Policy~\cite{DP:Chi:IJRR2025} was used as the imitation learning algorithm.
The state history length and action prediction horizon were set to 2 and 8, respectively.
The joint angles and tactile measurements were independently normalized to the range $[-1,1]$.
For the tactile input, normalization over all sensor cells yielded more stable training than per-cell normalization.

Data collection, policy training, and policy rollout were performed using the open-source software framework RoboManipBaselines~\cite{RMB:Murooka:Access2026}.





\subsection{Qualitative Evaluation of Learned Policies}

\begin{figure}[!htbp]
\centering
\includegraphics[width=0.85\textwidth]{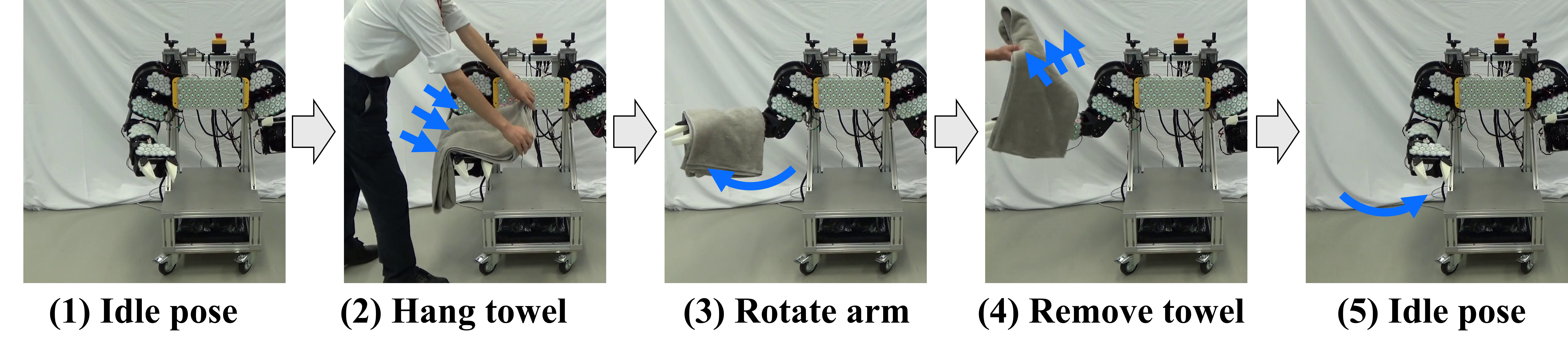}
\par\smallskip \scriptsize
(a) Towel Hanging
\par\medskip
\vspace{2mm}
\includegraphics[width=0.85\textwidth]{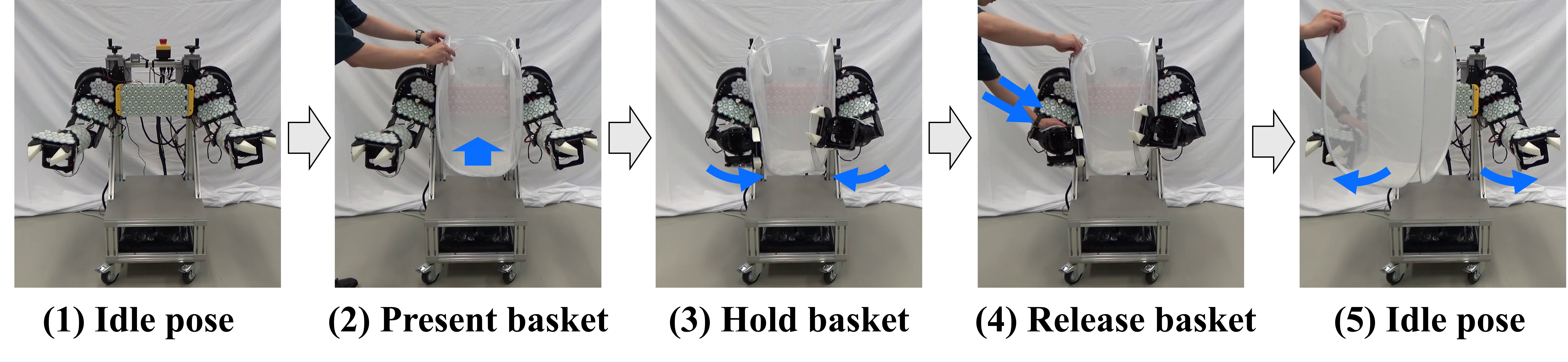}
\par\smallskip \scriptsize
(b) Basket Holding
\par\medskip
\vspace{2mm}
\includegraphics[width=0.85\textwidth]{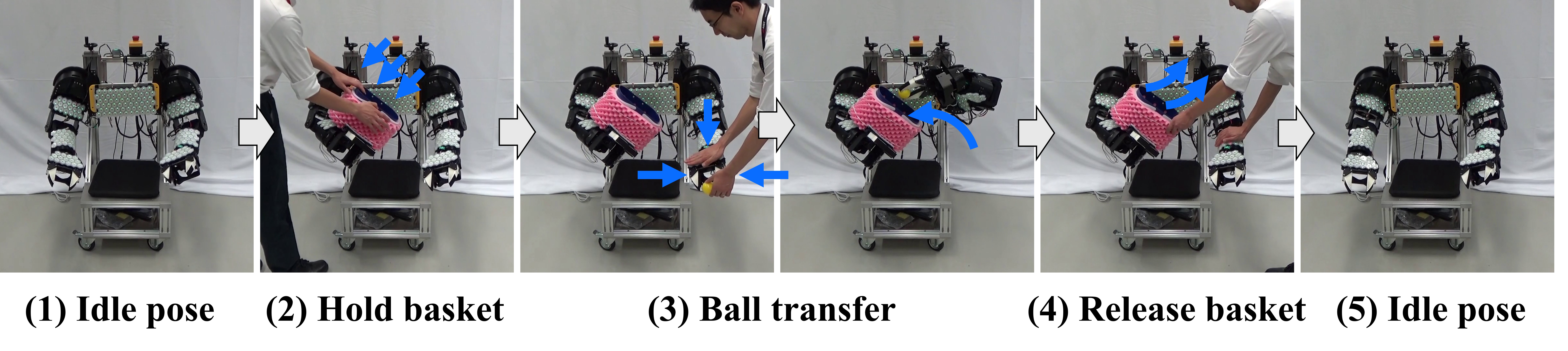}
\par\smallskip \scriptsize
(c) Ball Placing
\par\medskip
\vspace{2mm}
\includegraphics[width=0.85\textwidth]{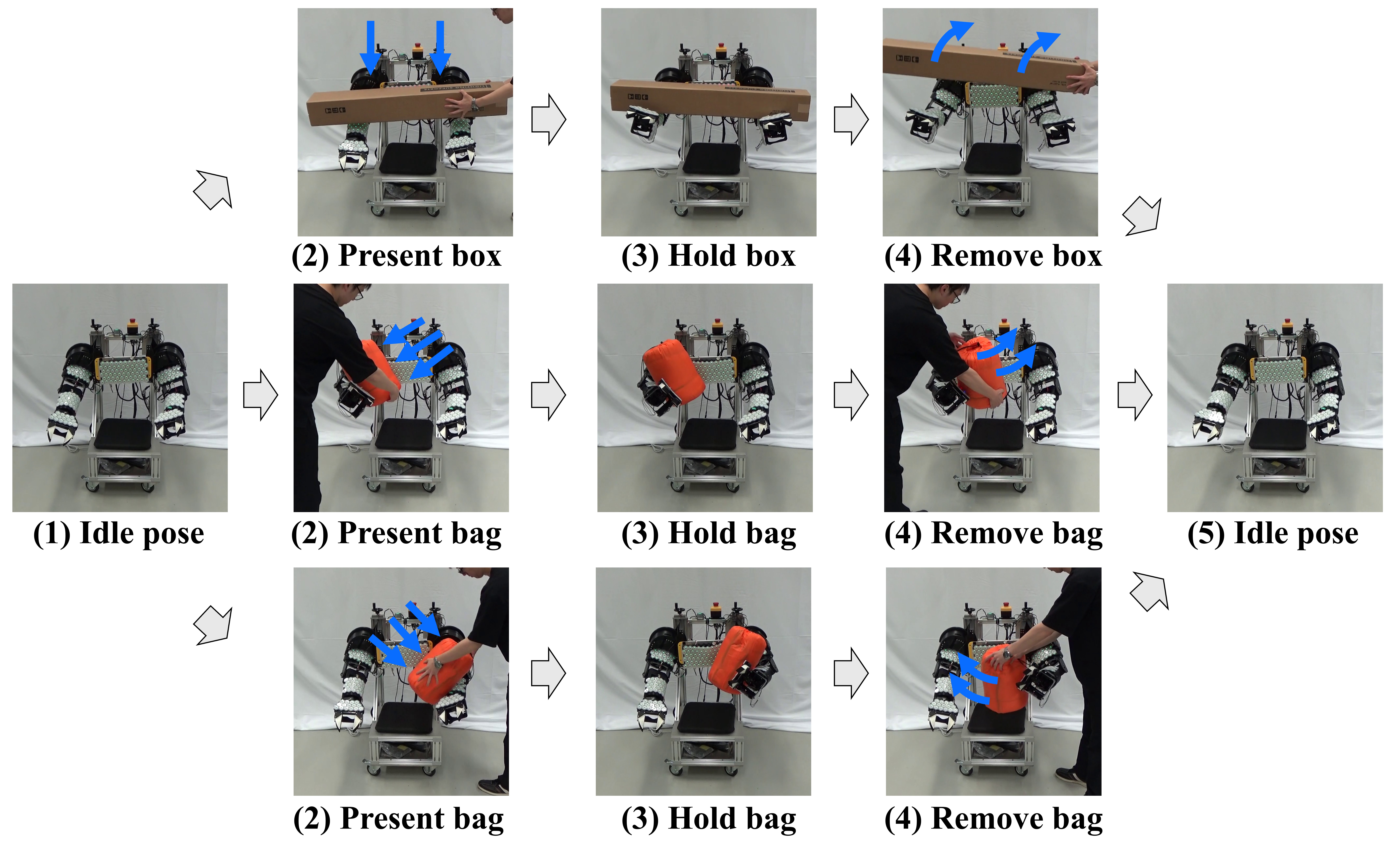}
\par\smallskip \scriptsize
(d) Adaptive Holding
\vspace{4mm}
\caption{Policy rollouts using the Isomorphic Robot.
\newline\footnotesize{
Each image corresponds to the same manipulation phase as in Figure~\ref{fig:teleop}.
}}
\label{fig:rollout}
\end{figure}

\begin{figure}[!htbp]
\centering
\includegraphics[width=0.9\textwidth]{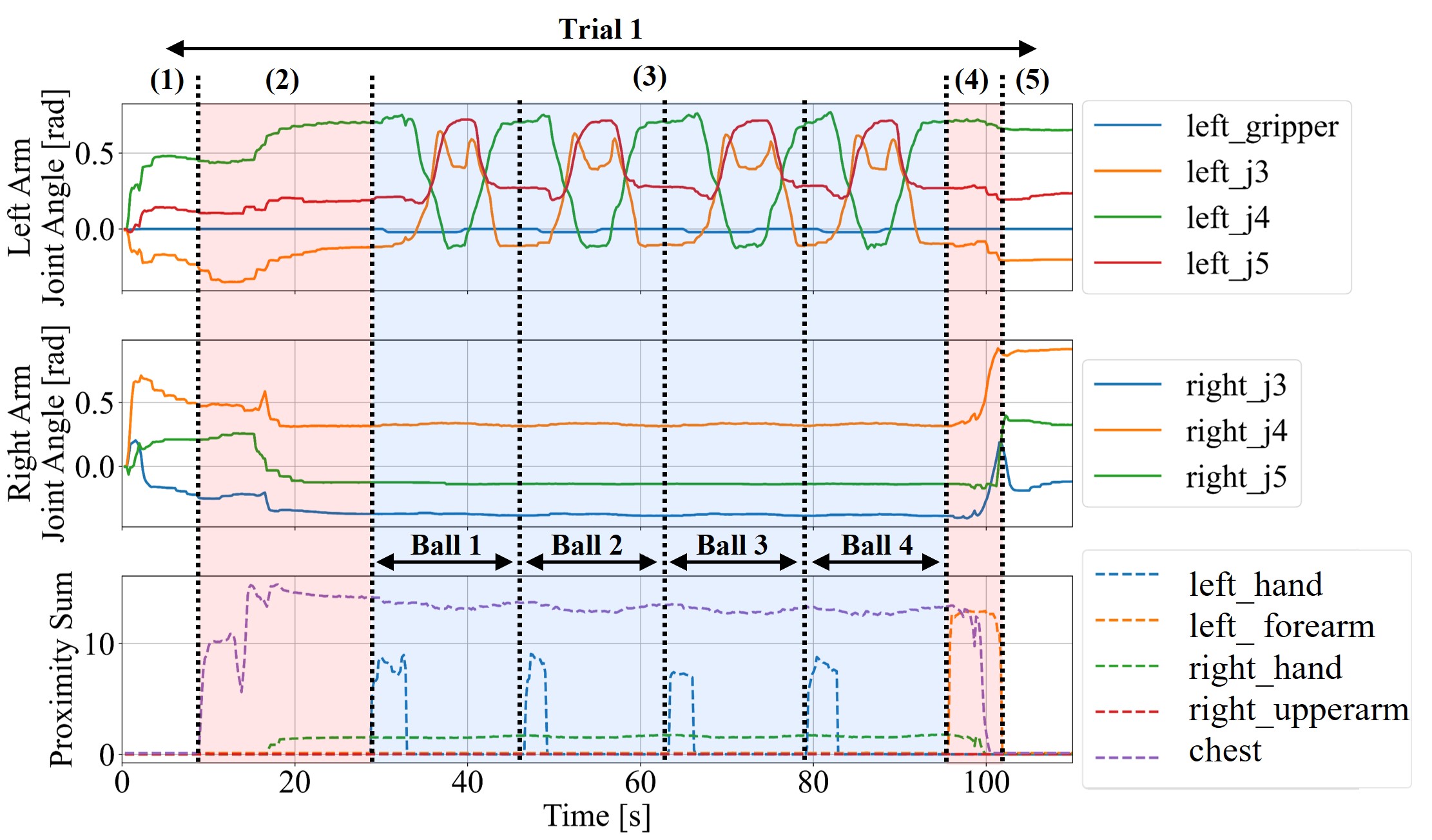}
\par\smallskip \scriptsize
(a) Ball Placing
\par\medskip
\vspace{2mm}
\includegraphics[width=0.9\textwidth]{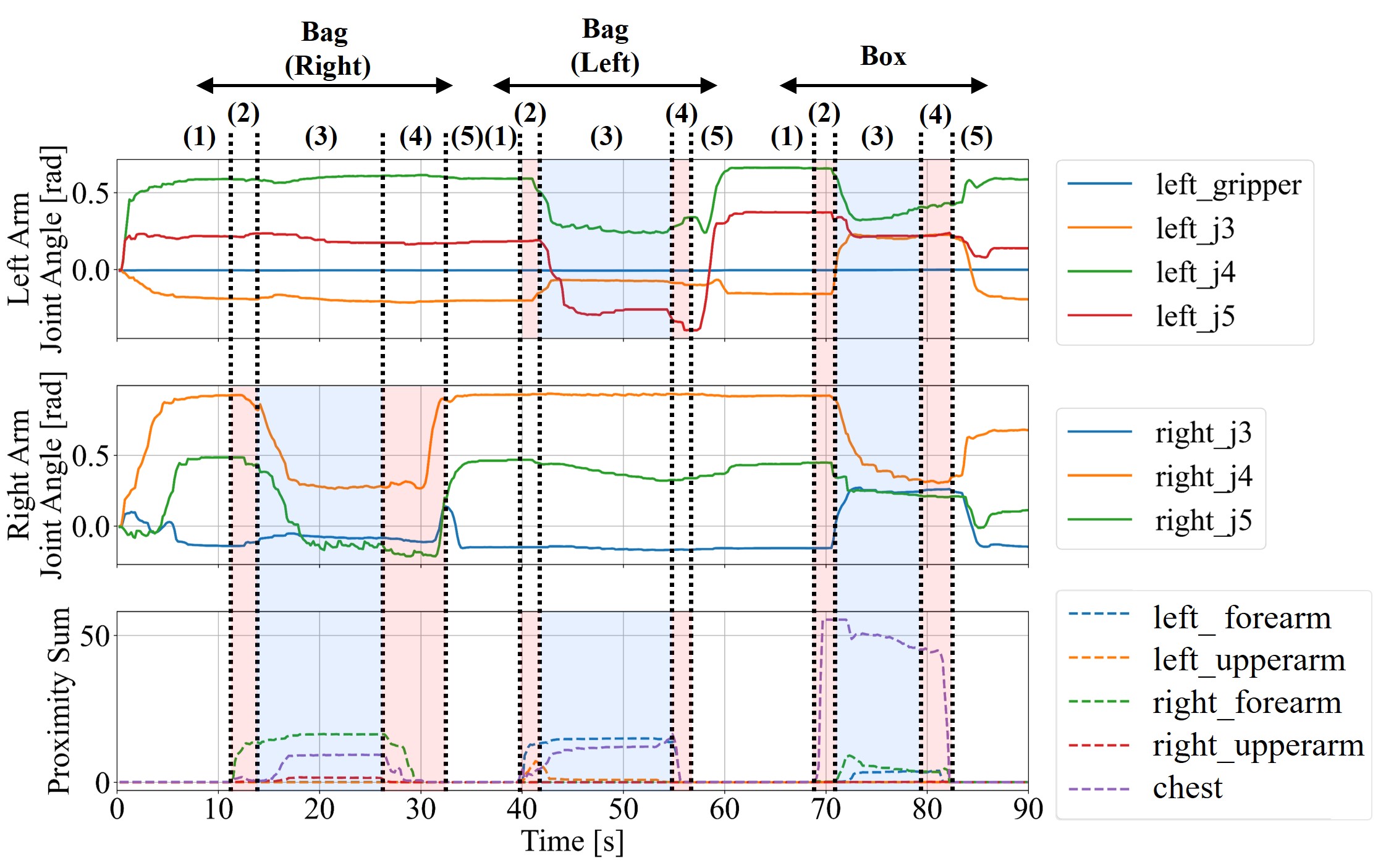}
\par\smallskip \scriptsize
(b) Adaptive Holding
\vspace{4mm}
\caption{Joint-angle and tactile-sensor data measured during policy rollouts with the Isomorphic Robot.
\newline\footnotesize{
The numbers above the plots correspond to the manipulation phases in Figure~\ref{fig:rollout}.
For clarity, only the same joint angles and tactile sensor patches as those in Figure~\ref{fig:teleop-graphs} are shown.
For each tactile sensor patch, the plotted value is the summed proximity measurement over all sensor cells.
Compared with the demonstrations shown in Figure~\ref{fig:teleop-graphs}, the number of ball presentations in the Ball Placing task and the presentation order of the objects in the Adaptive Holding task are different.
Nevertheless, the learned policies generate appropriate behaviors in response to body-surface contact events.
}}
\label{fig:rollout-graphs}
\end{figure}

\begin{figure}[!htbp]
\centering
\includegraphics[width=0.80\textwidth]{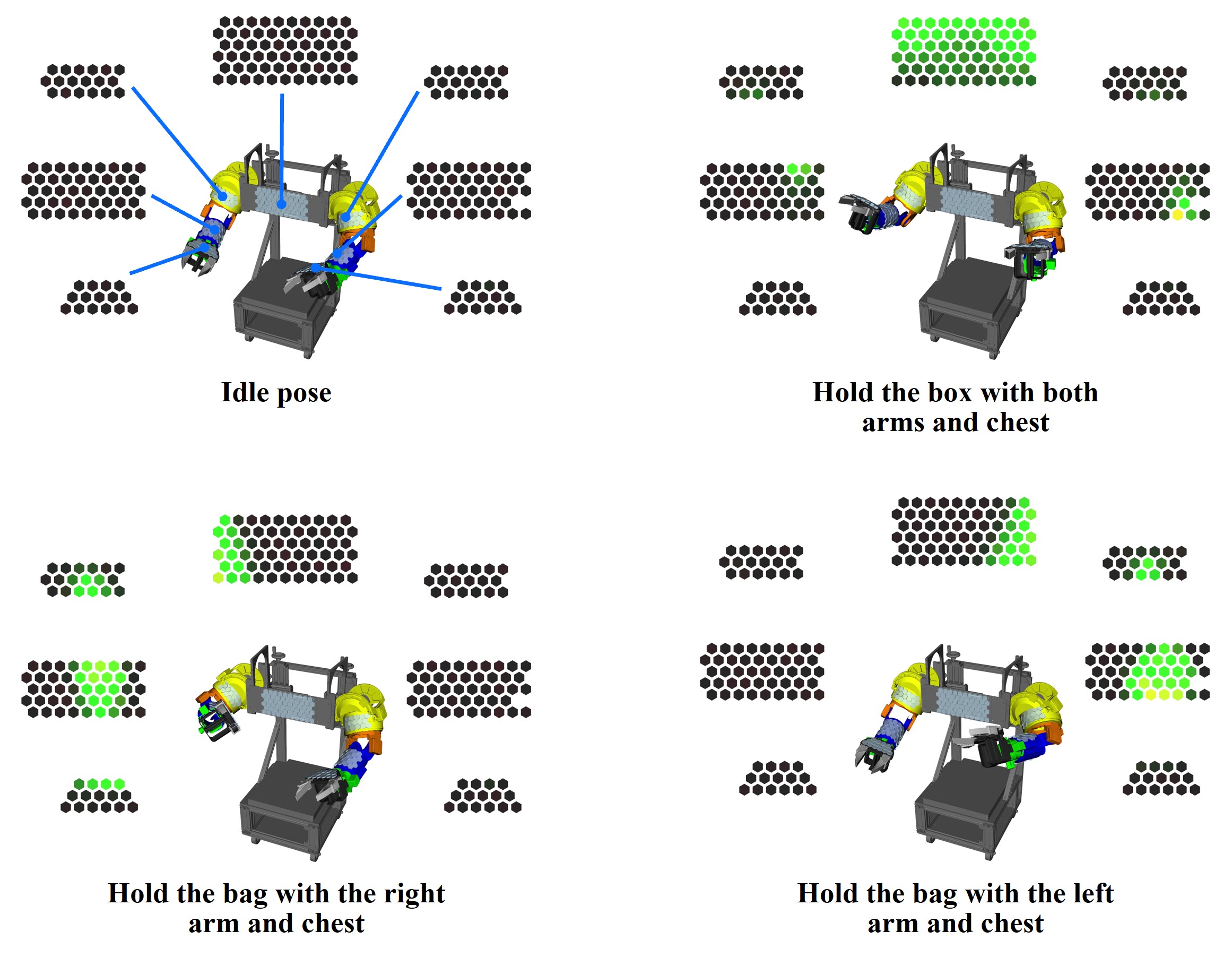}
\par\smallskip \scriptsize
(a) Demonstration collection with the Wearable Dual-Arm Device
\par\medskip
\vspace{4mm}
\includegraphics[width=0.80\textwidth]{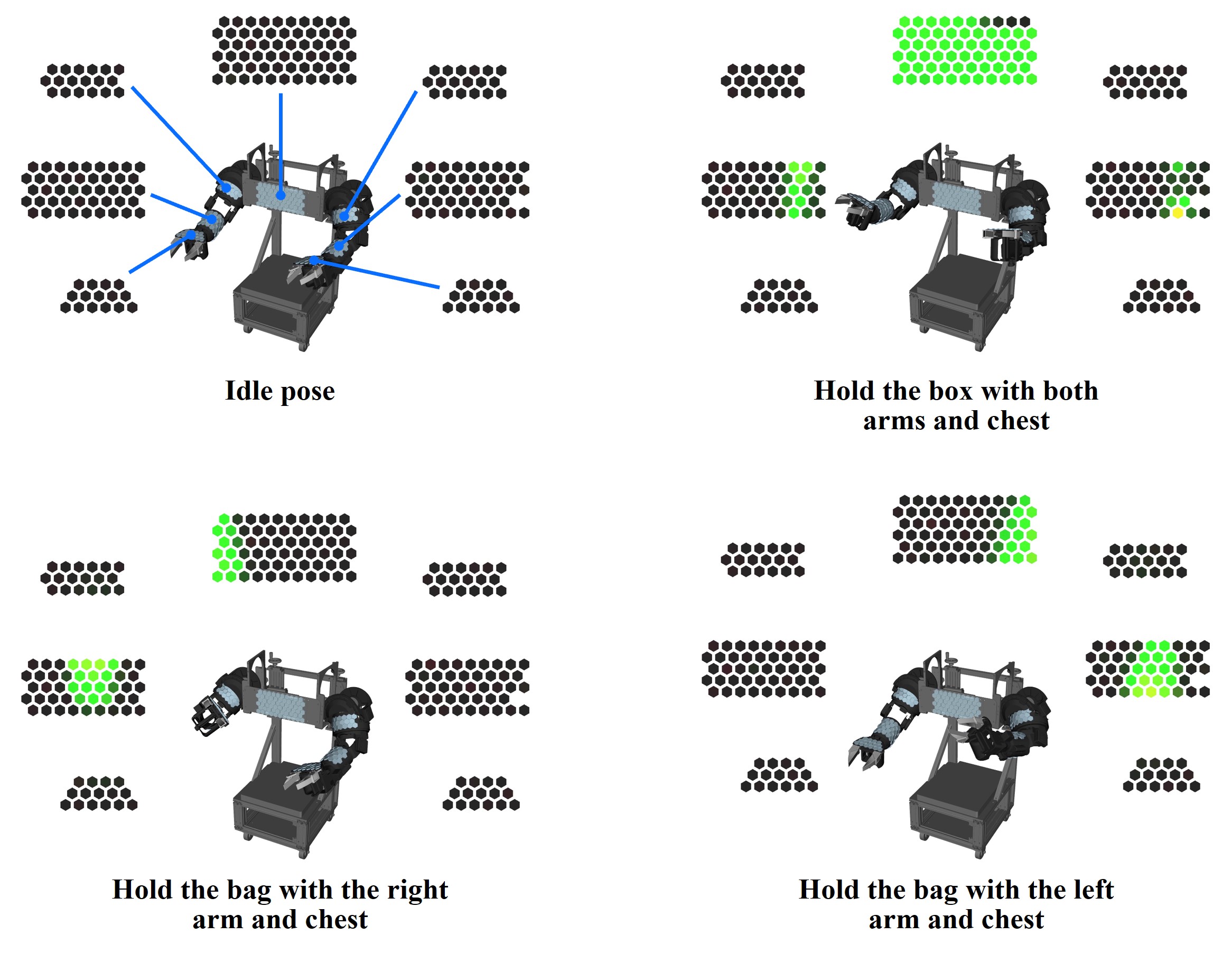}
\par\smallskip \scriptsize
(b) Policy rollout with the Isomorphic Robot
\vspace{4mm}
\caption{Visualization of tactile sensor measurements during the Adaptive Holding task.
Green indicates cells with an active proximity response, whereas yellow indicates cells with both active proximity and tactile responses.}
\label{fig:tactile-data}
\end{figure}

The learned policies were deployed on the Isomorphic Robot.
The purpose of this experiment is not to evaluate the imitation learning algorithm itself, but to verify that demonstration data collected with TWINS enable learning and execution of manipulation tasks involving body-surface contact.
Therefore, the evaluation focuses on qualitative observations of the robot behavior.

The Isomorphic Robot tracked the commanded joint trajectories with a mean absolute error of 0.94~deg over the 14 arm joints, with a 95th-percentile error of 3.43~deg.
The remaining error was mainly attributed to a tracking delay of approximately 0.4--0.6~s.
Despite these tracking limitations, the Isomorphic Robot successfully executed the contact-rich manipulation tasks using the learned policies.

Figure~\ref{fig:rollout} shows the execution results for the four tasks.
Figure~\ref{fig:rollout-graphs} presents examples of the joint-angle and tactile sensor data during the Ball Placing and Adaptive Holding tasks.
As shown in Figure~\ref{fig:rollout}, the learned policies successfully transitioned between manipulation phases according to body-surface contact events and reproduced the demonstrated behaviors in all tasks.
The plots in Figure~\ref{fig:rollout-graphs} further show that the joint motions changed in response to body-surface contact events, consistent with the demonstrations.

Figure~\ref{fig:tactile-data} visualizes the tactile sensor measurements during demonstration collection and policy rollout for the Adaptive Holding task.
The activation patterns of the tactile sensor patches are largely consistent for each holding posture, indicating that matching the joint configuration and tactile sensor layout between the Wearable Dual-Arm Device and the Isomorphic Robot enables effective transfer of body-surface contact from demonstrations to policy execution.




Although object presentation was manually performed under conditions that did not exactly match those during demonstration collection, the learned policies remained robust to the resulting variations in body-surface contact events.
Moreover, the policies successfully handled situations not included in the demonstrations.
For example, in the Towel Hanging task, the robot correctly rotated the corresponding arm when the order of towel presentation was reversed or when towels were simultaneously placed on both forearms.
In the Ball Placing task, the robot repeatedly placed balls into the basket in response to contact events, even when more balls were presented than in the demonstrations.
In the Adaptive Holding task, the robot selected the appropriate holding strategy even when the presentation order of the box and packed sleeping bag differed from that in the demonstrations.
These results indicate that the learned policies did not simply replay the demonstrations but instead learned to switch manipulation phases according to body-surface contact events.

The desired behaviors were successfully achieved in most trials across all tasks.
Occasional failures occurred in tasks involving holding an object between one arm and the chest, where the object was sometimes dropped.
This issue was substantially mitigated by attaching sponge padding to both the object and the chest surface, increasing the compliance of the contact interface.
Further improving the stability of holding motions may require explicitly controlling the joint torque during object holding.



\section{Conclusion}

This paper presented TWINS, a robot system for imitation learning of manipulation tasks involving body-surface contact.
TWINS consists of a Wearable Dual-Arm Device and an Isomorphic Robot that share the same joint configuration and external dimensions, enabling the collection of demonstration data integrating joint motions and tactile information.
Tactile sensors distributed over the robot body further enable manipulation using the arms and chest in addition to the grippers.

Using the Wearable Dual-Arm Device, demonstration data were collected for four representative manipulation tasks involving body-surface contact.
Imitation learning was performed using the collected data, and the learned policies were deployed on the Isomorphic Robot to execute the tasks.
These results demonstrate that TWINS can serve as a practical platform for collecting real-world demonstration data integrating joint motions and body-surface contact, and for applying such data to robot learning.

Future work includes extending TWINS to support mobile manipulation by allowing the operator to move during demonstration collection.
We also plan to improve the mechanism by incorporating human scapular motion, achieving closer anatomical fidelity, and introducing soft body surfaces to improve the stability and safety of body-surface contact.
Finally, by releasing the hardware design as open hardware, we aim to further develop TWINS as a data collection platform for robot learning with body-surface contact.




\backmatter

\section*{Declarations}

\bmhead{Funding}

This work was partially supported by JST CREST, Japan, under Grant Number
JPMJCR2553.

\bibliography{sn-bibliography}

@INPROCEEDINGS{HOMIE:Ben:RSS2025,
    AUTHOR    = {Qingwei Ben AND Feiyu Jia AND Jia Zeng AND Junting Dong AND Dahua Lin AND Jiangmiao Pang},
    TITLE     = {{HOMIE}: Humanoid Loco-Manipulation with Isomorphic Exoskeleton Cockpit},
    BOOKTITLE = {Robotics: Science and Systems},
    YEAR      = {2025},
}

@article{RMB:Murooka:Access2026,
  author={Murooka, Masaki and Motoda, Tomohiro and Nakajo, Ryoichi and Oh, Hanbit and Makihara, Koshi and Shirai, Keisuke and Ogata, Tetsuya and Domae, Yukiyasu},
  journal={IEEE Access},
  title={{RoboManipBaselines}: A Unified Framework for Imitation Learning in Robotic Manipulation Across Real and Simulation Environments},
  year={2026},
  volume={14},
  number={},
  pages={97896-97909},
}

@article{DP:Chi:IJRR2025,
author = {Cheng Chi and Zhenjia Xu and Siyuan Feng and Eric Cousineau and Yilun Du and Benjamin Burchfiel and Russ Tedrake and Shuran Song},
title ={Diffusion policy: Visuomotor policy learning via action diffusion},
journal = {The International Journal of Robotics Research},
volume = {44},
number = {10-11},
pages = {1684-1704},
year = {2025},
}

@ARTICLE{RobotSkin:Cheng:IEEE2019,
  author={Cheng, Gordon and Dean-Leon, Emmanuel and Bergner, Florian and Rogelio Guadarrama Olvera, Julio and Leboutet, Quentin and Mittendorfer, Philipp},
  journal={Proceedings of the IEEE},
  title={A comprehensive realization of robot skin: sensors, sensing, control, and applications},
  year={2019},
  volume={107},
  number={10},
  pages={2034-2051},
}

@INPROCEEDINGS{HRP2Kai:Kaneko:Humanoids2015,
  author={Kaneko, Kenji and Morisawa, Mitsuharu and Kajita, Shuuji and Nakaoka, Shin'ichiro and Sakaguchi, Takeshi and Cisneros, Rafael and Kanehiro, Fumio},
  booktitle={International Conference on Humanoid Robots}, 
  title={Humanoid robot HRP-2Kai — Improvement of HRP-2 towards disaster response tasks}, 
  year={2015},
  volume={},
  number={},
  pages={132-139},
  }

@inproceedings{ActionSense:DelPreto:NeurIPS2022,
title={{ActionSense}: A Multimodal Dataset and Recording Framework for Human Activities Using Wearable Sensors in a Kitchen Environment},
author={Joseph DelPreto and Chao Liu and Yiyue Luo and Michael Foshey and Yunzhu Li and Antonio Torralba and Wojciech Matusik and Daniela Rus},
booktitle={Neural Information Processing Systems Track on Datasets and Benchmarks},
year={2022}
}

@INPROCEEDINGS{CHILD:Myers:Humanoids2025,
  author={Myers, Noboru and Kwon, Obin and Yamsani, Sankalp and Kim, Joohyung},
  booktitle={International Conference on Humanoid Robots}, 
  title={{CHILD (Controller for Humanoid Imitation and Live Demonstration)}: A Whole-Body Humanoid Teleoperation System}, 
  year={2025},
  volume={},
  number={},
  pages={1-6},
  }

@ARTICLE{TABLIS:Ishiguro:RAL2020,
  author={Ishiguro, Yasuhiro and Makabe, Tasuku and Nagamatsu, Yuya and Kojio, Yuta and Kojima, Kunio and Sugai, Fumihito and Kakiuchi, Yohei and Okada, Kei and Inaba, Masayuki},
  journal={IEEE Robotics and Automation Letters}, 
  title={Bilateral Humanoid Teleoperation System Using Whole-Body Exoskeleton Cockpit TABLIS}, 
  year={2020},
  volume={5},
  number={4},
  pages={6419-6426},
}

@ARTICLE{MasterRobot:Kim:RAL2023,
  author={Kim, Heecheol and Ohmura, Yoshiyuki and Nagakubo, Akihiko and Kuniyoshi, Yasuo},
  journal={IEEE Robotics and Automation Letters}, 
  title={Training Robots Without Robots: Deep Imitation Learning for Master-to-Robot Policy Transfer}, 
  year={2023},
  volume={8},
  number={5},
  pages={2906-2913},
}

@inproceedings{UMI:Chi:RSS2024,
	title={Universal Manipulation Interface: In-The-Wild Robot Teaching Without In-The-Wild Robots},
	author={Chi, Cheng and Xu, Zhenjia and Pan, Chuer and Cousineau, Eric and Burchfiel, Benjamin and Feng, Siyuan and Tedrake, Russ and Song, Shuran},
	booktitle={Robotics: Science and Systems},
	year={2024}
}

@article{MEVION:Kawaharazuka:arXiv2026,
  title={{MEVION}: Low-Cost Open-Source Data Collection System for Powerful and High-Speed Dual-Arm Manipulation},
  author={Kawaharazuka, Kento and Obinata, Yoshiki and Ishida, Hirokazu and Oh, Jihoon and Suzuki, Temma and Inoue, Shintaro and Yoneda, Keita and Iwata, Ayumu and Okada, Kei},
  journal={arXiv preprint arXiv:2607.17970},
  year={2026}
}

@ARTICLE{TeleopHumanoid:Penco:RAM2019,
  author={Penco, Luigi and Scianca, Nicola and Modugno, Valerio and Lanari, Leonardo and Oriolo, Giuseppe and Ivaldi, Serena},
  journal={IEEE Robotics \& Automation Magazine}, 
  title={A Multimode Teleoperation Framework for Humanoid Loco-Manipulation: An Application for the iCub Robot}, 
  year={2019},
  volume={26},
  number={4},
  pages={73-82},
}

@inproceedings{OmniH2O:He:CoRL2024,
title={{OmniH2O}: Universal and Dexterous Human-to-Humanoid Whole-Body Teleoperation and Learning},
author={Tairan He and Zhengyi Luo and Xialin He and Wenli Xiao and Chong Zhang and Weinan Zhang and Kris M. Kitani and Changliu Liu and Guanya Shi},
booktitle={Conference on Robot Learning},
year={2024},
}

@ARTICLE{TACT:Murooka:RAL2025,
  author={Murooka, Masaki and Hoshi, Takahiro and Fukumitsu, Kensuke and Masuda, Shimpei and Hamze, Marwan and Sasaki, Tomoya and Morisawa, Mitsuharu and Yoshida, Eiichi},
  journal={IEEE Robotics and Automation Letters}, 
  title={{TACT}: Humanoid Whole-Body Contact Manipulation Through Deep Imitation Learning With Tactile Modality}, 
  year={2025},
  volume={10},
  number={8},
  pages={7819-7826},
}

@ARTICLE{HumanoidRetarget:Ayusawa:TRO2017,
  author={Ayusawa, Ko and Yoshida, Eiichi},
  journal={IEEE Transactions on Robotics}, 
  title={Motion Retargeting for Humanoid Robots Based on Simultaneous Morphing Parameter Identification and Motion Optimization}, 
  year={2017},
  volume={33},
  number={6},
  pages={1343-1357},
}

@INPROCEEDINGS{ALOHA:Zhao:RSS2023,
    AUTHOR    = {Tony Z. Zhao AND Vikash Kumar AND Sergey Levine AND Chelsea Finn},
    TITLE     = {Learning Fine-Grained Bimanual Manipulation with Low-Cost Hardware},
    BOOKTITLE = {Robotics: Science and Systems},
    YEAR      = {2023},
}

@inproceedings{MobileALOHA:Fu:CoRL2024,
title={Mobile {ALOHA}: Learning Bimanual Mobile Manipulation using Low-Cost Whole-Body Teleoperation},
author={Zipeng Fu and Tony Z. Zhao and Chelsea Finn},
booktitle={Conference on Robot Learning},
year={2024},
}

@article{HumanoidExo:Zhong:arXiv2025,
  title={{HumanoidExo}: Scalable Whole-Body Humanoid Manipulation via Wearable Exoskeleton},
  author={Zhong, Rui and Sun, Yizhe and Wen, Junjie and Li, Jinming and Cheng, Chuang and Dai, Wei and Zeng, Zhiwen and Lu, Huimin and Zhu, Yichen and Xu, Yi},
  journal={arXiv preprint arXiv:2510.03022},
  year={2025}
}

@article{TeleopReview:Liang:Neuro2025,
title = {Tremor suppression for master-slave teleoperated robot based on machine learning: A review},
journal = {Neurocomputing},
volume = {623},
pages = {129421},
year = {2025},
issn = {0925-2312},
author = {Ke Liang and Yue Su and Gang Du and Chun Ma and Mantian Li and Mingzhang Pan}
}

@article{ConRichSurvey:Tsuji:IJRR2025,
  title={A survey on imitation learning for contact-rich tasks in robotics},
  author={Tsuji, Toshiaki and Kato, Yasuhiro and Solak, Gokhan and Zhang, Heng and Petri{\v{c}}, Tadej and Nori, Francesco and Ajoudani, Arash},
  journal={International Journal of Robotics Research},
  pages={02783649261417694},
  year={2025},
}

@INPROCEEDINGS{OpenX:OpenX:ICRA2024,
  author={{Open X-Embodiment Collaboration}},
  booktitle={IEEE International Conference on Robotics and Automation},
  title={{Open X-Embodiment}: Robotic learning datasets and {RT-X} models},
  year={2024},
  volume={},
  number={},
  pages={6892-6903},
  }

@INPROCEEDINGS{RH20T:Fang:ICRA2024,
  author={Fang, Hao-Shu and Fang, Hongjie and Tang, Zhenyu and Liu, Jirong and Wang, Chenxi and Wang, Junbo and Zhu, Haoyi and Lu, Cewu},
  booktitle={IEEE International Conference on Robotics and Automation},
  title={{RH20T}: A Comprehensive Robotic Dataset for Learning Diverse Skills in One-Shot},
  year={2024},
  volume={},
  number={},
  pages={653-660},
  }

@article{ILSurvey:Osa:FTR2018,
  title={An algorithmic perspective on imitation learning},
  author={Osa, Takayuki and Pajarinen, Joni and Neumann, Gerhard and Bagnell, J Andrew and Abbeel, Pieter and Peters, Jan},
  journal={Foundations and Trends{\textregistered} in Robotics},
  volume={7},
  number={1-2},
  pages={1--179},
  year={2018},
  publisher={Emerald Publishing Limited}
}

@ARTICLE{EmbodiedIntelligenceReview:Zhang:FRAI2025,
AUTHOR={Zhang, Yunwei  and Tian, Jing  and Xiong, Qiaochu },
TITLE={A review of embodied intelligence systems: a three-layer framework integrating multimodal perception, world modeling, and structured strategies},
JOURNAL={Frontiers in Robotics and AI},
VOLUME={12},
YEAR={2025},
ISSN={2296-9144}}

\end{document}